\documentclass[11pt]{article}

\usepackage{PRIMEarxiv}

\usepackage[T1]{fontenc}
\usepackage[utf8]{inputenc}
\usepackage{amsmath,amssymb}
\usepackage{graphicx}
\usepackage{booktabs}
\usepackage{multirow}
\usepackage{array}
\usepackage{longtable}
\usepackage{xcolor}
\usepackage{tcolorbox}
\usepackage{listings}
\usepackage{upquote} 
\usepackage{enumitem}
\usepackage{tikz}
\usetikzlibrary{arrows.meta, positioning, shapes.geometric, fit, shadows, calc}
\usepackage{caption}
\usepackage{pifont}
\usepackage[strings]{underscore} 
\usepackage{adjustbox} 
\usepackage{microtype} 
\usepackage{placeins} 
\usepackage{needspace} 
\usepackage{hyperref}

\usetikzlibrary{backgrounds}
\definecolor{akslate}{RGB}{71,85,105}
\definecolor{akblue}{RGB}{37,99,235}
\definecolor{akteal}{RGB}{13,148,136}
\definecolor{akgreen}{RGB}{21,128,61}
\definecolor{akviolet}{RGB}{109,40,217}
\definecolor{akamber}{RGB}{180,83,9}
\definecolor{akgray}{RGB}{100,116,139}
\definecolor{akink}{RGB}{51,65,85}
\tikzset{
  akstep/.style={rectangle, rounded corners=2pt, thin, minimum height=0.42cm, align=flush center,
                 inner sep=2.6pt, fill=white, text width=3.3cm, text=akink,
                 font=\tiny},
  akmini/.style={-{Stealth[length=1.2mm]}, thin},
  akmain/.style={-{Stealth[length=2.4mm]}, very thick, akink!85},
  akbranch/.style={-{Stealth[length=1.8mm]}, dashed, thick},
  aklbl/.style={font=\scriptsize\itshape, align=center, text=akink!80},
  akbadge/.style={circle, thick, fill=white, minimum size=4.6mm, inner sep=0pt, font=\bfseries\scriptsize},
}
\newcommand{\akcard}[9]{%
  \coordinate (#1-top) at (#2,#3);
  \foreach \akstep [count=\akn] in {#9}{%
    \ifnum\akn=1
      \node[akstep, draw=#4!45, anchor=north] (#1-s\akn) at ($(#1-top)+(0,-0.78)$) {\akstep};
    \else
      \pgfmathtruncatemacro{\akprev}{\akn-1}%
      \node[akstep, draw=#4!45, below=1.6mm of #1-s\akprev] (#1-s\akn) {\akstep};
      \draw[akmini, #4!70] (#1-s\akprev.south) -- (#1-s\akn.north);
    \fi
    \xdef\aklast{\akn}%
  }%
  \coordinate (#1-end) at ($(#1-s\aklast.south)+(0,-0.22)$);
  \coordinate (#1-min) at ($(#1-top)+(0,-#7)$);
  \begin{scope}[on background layer]
    \node[fit=(#1-top)(#1-end)(#1-min)(#1-s1.west)(#1-s1.east), inner xsep=0.2cm, inner ysep=0pt,
          rectangle, rounded corners=6pt, fill=akink, opacity=0.12, xshift=1.6pt, yshift=-1.6pt] {};
    \node[fit=(#1-top)(#1-end)(#1-min)(#1-s1.west)(#1-s1.east), inner xsep=0.2cm, inner ysep=0pt,
          rectangle, rounded corners=6pt, fill=#4!5] (#1) {};
    \begin{scope}
      \clip[rounded corners=6pt] (#1.south west) rectangle (#1.north east);
      \fill[#4] (#1.north west) rectangle ([yshift=-0.6cm]#1.north east);
    \end{scope}
    \draw[#4, thick, rounded corners=6pt, #8] (#1.south west) rectangle (#1.north east);
  \end{scope}
  \node[font=\bfseries\small, text=white] at ([yshift=-0.31cm]#1.north) {#5};
  \node[akbadge, draw=#4, text=#4] at ([xshift=-3pt,yshift=3pt]#1.north west) {#6};
}

\newcommand{\ak}{\textit{AuditKIT}}

\newcommand{\code}[1]{\texttt{#1}}

\tcbuselibrary{skins,breakable}
\newtcolorbox{takeaway}{
  colback=gray!8, colframe=black, boxrule=0.5pt,
  fonttitle=\bfseries, coltitle=white, colbacktitle=black,
  title=Takeaway, breakable
}

\lstdefinestyle{code}{
  basicstyle=\ttfamily\small,
  breaklines=true,
  frame=single,
  backgroundcolor=\color{gray!5},
  columns=fullflexible,
  keywordstyle=\color{blue!70!black}\bfseries,
  commentstyle=\color{gray!60!black}\itshape,
  stringstyle=\color{green!40!black},
  showstringspaces=false,
  upquote=true,
  xleftmargin=3.4pt, xrightmargin=3.4pt, framesep=3pt,
  aboveskip=8pt, belowskip=8pt,
  basicstyle=\ttfamily\footnotesize
}
\title{\includegraphics[width=0.9\textwidth]{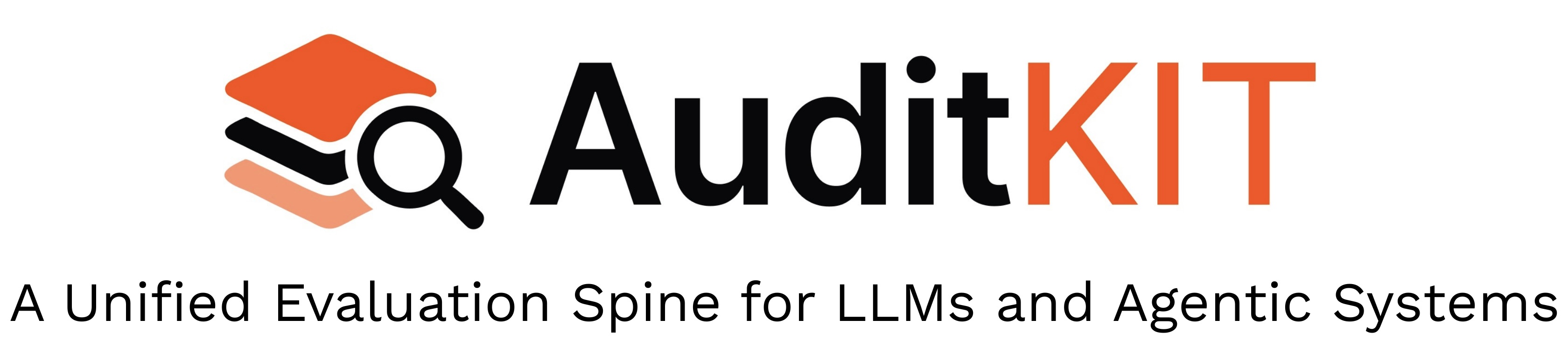}}

\author{
  Shreeyans Arora\textsuperscript{*\dag}, Utsav Avaiya\textsuperscript{*\dag}, \\
  Zera Lyngkhoi, Hem Gosalia, Ram Mohan Rao Kadiyala, \\
  Vinay Kumar Sankarapu, Pratinav Seth \\
  \affiliation{Lexsi Labs} \\
  \texttt{pratinav.seth@lexsi.ai}
}

\date{}

\setabstract{%
Evaluating language models and agents requires combining benchmark accuracy, judge-based assessment, runtime performance, tool-use behavior, and retrieval quality. Differences in interfaces, metric semantics, and trace formats complicate these comparisons and can hide failures in agentic workflows. We present AuditKIT, a source-available evaluation library that records these measurements in one result representation, with direction-aware metric contracts, configuration fingerprints, and extensible metrics and backends. For agents, it provides turn-aware tool-call matching, separates parallel from sequential execution, and distinguishes verified task outcomes from trajectory-level diagnostics. Missing or unobservable evidence is marked unscored rather than counted as failure. For retrieval-augmented generation, it combines ranked retrieval metrics, judged grounding, and deterministic evidence checks. Seven case studies spanning model comparison, serving runtimes, retrieval, tool calling, and interoperability show how answer extraction, runtime behavior, and metric selection change evaluation outcomes.}

\setkeywords{LLM Evaluation, Agent Evaluation, Tool Use, Retrieval-Augmented Generation, LLM-as-Judge, Model Comparison, Reproducibility}

\runningtitle{\ak: Evaluation for Language Models and Agents}

\begin{document}

\maketitle

{\renewcommand{\thefootnote}{\fnsymbol{footnote}}\footnotetext[1]{\vspace{0.5em} Equal contribution}\footnotetext[2]{Work done while at Lexsi Labs.}}

\section{Introduction}
\label{sec:intro}

Evaluating a language model involves more than a score on one benchmark. Different applications call for task accuracy, generation quality, runtime efficiency, or safety, and each uses its own measurements. Agents add further requirements~\cite{yao2023react,schick2023toolformer,lewis2020rag}. Their behavior depends on tool selection, argument correctness, execution order, retrieved contexts, and changes to the external environment, not only on generated text. A model can write a plausible final answer while the underlying task fails, and an agent can reach the correct outcome by a path that differs from a predefined reference.

Existing frameworks address many of these needs, but their interfaces, metric definitions, and result formats differ. \texttt{lm-evaluation-harness}~\cite{lmevalharness} and HELM~\cite{helm} standardize benchmark evaluation; LangSmith~\cite{langsmith}, DeepEval~\cite{deepeval}, and Braintrust~\cite{braintrust} support application-level evaluation and experiment workflows; Promptfoo~\cite{promptfoo} and PyRIT~\cite{pyrit} support red teaming. Combining their outputs requires attention to how each metric is computed, what evidence is available, and whether results from different configurations are comparable.

Several practical problems follow. Metrics differ in optimization direction, so a higher value is better for some and worse for others. Several tools store direction for some metrics (Table~\ref{tab:comparison}), but among the tools reviewed (versions in Section~\ref{sec:positioning}) we found none that declares it in the metric definition and applies it in every model comparison, which matters once accuracy, latency, error rate, and cost are ranked together. Tool calls scored apart from the turns that issued them hide the difference between sequential and parallel execution. A judge's ``looks complete'' can stand in for task completion when the final state is missing or contradicts it. Trace fields the recorder never captured can be read as empty, which turns missing evidence into a measured zero.

We present \ak, a source-available evaluation library for language models and agents. It provides one interface for benchmark evaluation, model-based judging, performance measurement, tool-use assessment, and retrieval-augmented generation, and it marks unavailable evidence as unscored so that measurements that could not be made are not counted as failures. The contributions below list the components.

The paper describes the architecture, metric interfaces, agent evaluation methodology, and interoperability with external artifacts. Seven case studies show how answer extraction, serving behavior, retrieval metrics, and trace structure change the interpretation of evaluation results. They do not establish model rankings; they show the distinctions \ak\ is designed to preserve.

\needspace{6\baselineskip}
\paragraph{Contributions.}
\begin{itemize}[leftmargin=*, itemsep=2pt, topsep=2pt]
  \item \emph{Evaluation interface.} \code{evaluate()} turns a dataset, adapter, model, optional annotator, and metrics into one \code{RunResult}, consumed by \code{compare\_models()} for pairwise and $N$-model comparison (Section~\ref{sec:architecture}).
  \item \emph{Direction-aware comparison and provenance.} Every metric declares \code{MAXIMIZE} or \code{MINIMIZE}, enforced at class definition and used by every comparison (Section~\ref{sec:design}); a SHA-256 fingerprint over model, data, adapter, metrics, and settings serves as cache key and comparison key.
  \item \emph{Extensible metrics and backends.} Built-in and custom metrics and model backends share one evaluation contract (Section~\ref{sec:metrics}).
  \item \emph{Agent and RAG evaluation.} Turn-aware tool-call metrics with bipartite matching, parallel-call scoring, multi-route references, an episode layer with verified outcomes and trial reliability, ranked retrieval, judged grounding, and deterministic RAG evidence checks (Section~\ref{sec:agentic}).
  \item \emph{Interoperability.} Read-only importers for AgentTune~\cite{agenttune}, AlignTune~\cite{aligntune}, SafeTune~\cite{safetune}, CuratorKIT~\cite{curatorkit}, and CircuitKIT~\cite{circuitkit}; zero required dependencies in the core.
\end{itemize}

\FloatBarrier
\section{Background and Related Work}
\label{sec:related}

\subsection{Evaluation frameworks}
\label{sec:related-tooling-today}

Benchmark harnesses such as \texttt{lm-eval-harness}~\cite{lmevalharness}, HELM~\cite{helm}, and OpenAI Evals~\cite{openaievals} give standardized task coverage. LLM-as-judge evaluation is served by code-first libraries (DeepEval~\cite{deepeval}, RAGAS~\cite{ragas}, TruLens~\cite{trulens}, Inspect~\cite{inspect}) and hosted platforms (LangSmith~\cite{langsmith}, Braintrust~\cite{braintrust}, Arize Phoenix~\cite{arizephoenix}, W\&B Weave~\cite{wandbweave}, MLflow~\cite{mlflow}). Red-teaming suites such as Promptfoo~\cite{promptfoo}, PyRIT~\cite{pyrit}, DeepTeam~\cite{deepteam}, and Giskard~\cite{giskard} probe for vulnerabilities. Performance is usually measured by separate instrumentation. Existing frameworks cover substantial parts of this workflow but differ in how they represent metrics, traces, execution configurations, and comparison results. \ak\ addresses red teaming only at a basic level (Section~\ref{sec:limitations}).

\subsection{Agent and retrieval evaluation}
\label{sec:related-agents}

Tool-use evaluation is mostly benchmark-specific. BFCL~\cite{patil2025bfcl} checks calls against accepted-value lists, including parallel and irrelevance categories; API-Bank~\cite{li2023apibank} and ToolLLM~\cite{qin2024toolllm} evaluate API selection over large inventories; AgentBench~\cite{liu2024agentbench} covers interactive environments; and $\tau$-bench~\cite{yao2024taubench} compares the final database state to a goal state and introduces the pass$^k$ reliability measure. Libraries add evaluators on top: DeepEval, Inspect, LangChain's \texttt{agentevals}~\cite{agentevals}, and Promptfoo's trajectory assertions for agents; RAGAS, TruLens, ARES~\cite{saadfalcon2024ares}, and RAGBench~\cite{friel2024ragbench} for RAG, mostly through judged faithfulness and context metrics; ARES and RAGBench propose judges and study judge-human agreement, documenting cases where judged scores diverge from human labels.

Reviewing the scoring code of RAGAS, DeepEval, TruLens, Phoenix, BFCL, \texttt{agentevals}, Inspect, and $\tau$-bench, we found three recurring choices that \ak\ departs from: calls are paired greedily (BFCL, \texttt{agentevals}, DeepEval) or sorted and compared as sets (RAGAS), where \ak\ uses a maximum bipartite matching~\cite{kuhn1955hungarian}; calls are flattened across turns, and none separates serialized from batched calls with the same call-level score (BFCL's \code{parallel} categories test parallel calling, but as an all-or-nothing set); and RAG average precision often divides by the relevant items retrieved, hiding misses. Benchmarks are complementary: BFCL loads directly as an \ak\ dataset, and $\tau$-bench's state-based success and pass$^k$ are the model for \ak's oracles and reliability report. The structural scores diagnose adherence to a reference execution structure; they do not independently establish task success.

\subsection{AuditKIT in relation to existing frameworks}
\label{sec:positioning}

Table~\ref{tab:comparison} compares \ak\ with ten tools across benchmarks, judging, performance, comparison semantics, provenance, deployment, agents, RAG, and red teaming. Versions checked: \texttt{lm-evaluation-harness} v0.4.12, HELM (\texttt{crfm-helm}) v0.5.16, DeepEval v4.2.0, Inspect AI v0.3.260, Arize Phoenix v17.2.0, and Promptfoo v0.122.0 on 2026-08-27, and PyRIT v1.1.0 on 2026-10-06; LangSmith and Braintrust are hosted platforms without public versions, and OpenAI Evals is an unversioned reference repository. We checked direction-awareness and provenance against public documentation; other feature entries summarize public product descriptions and are intended as positioning. A mark means we did not find the capability in that version's documentation, not that it cannot be built from other features.

Several marks in Table~\ref{tab:comparison} need qualification. \texttt{lm-eval-harness} stores per-metric \code{higher\_is\_better} metadata, but we found no $N$-model leaderboard API that uses it for winner selection. HELM's Mean Win Rate is direction-correct but a fixed aggregate, not a per-metric contract for custom metrics. LangSmith lets users mark a feedback key higher-is-better in experiment comparison, outside a metric-definition contract. DeepEval supports higher- and lower-is-better metrics, but not as a general contract for cross-model ranking. Braintrust compares latency, cost, errors, and load with fixed semantics, not for arbitrary custom scorers. We found no direction-aware $N$-model winner selection in OpenAI Evals, and Inspect has structured, reproducible logs but no metric-direction contract tied to winner selection. Phoenix evaluates traced data but has no direction-aware winner selection or run fingerprint. Promptfoo compares providers and caches individual responses, but its latency and cost assertions are thresholds, not direction-aware winner selection; its \code{tool-call-f1} compares tool \emph{names}, and its \code{trajectory:*} assertions check tools used, arguments, order, and an LLM-judged goal. PyRIT~\cite{pyrit} scores attack success (true/false or float scales) and stores every prompt and score in a SQL memory, but has no cross-model winner selection or run-level fingerprint.

\begin{table}[t]
\centering
\caption{Feature coverage of AuditKIT and selected evaluation tools.}
\label{tab:comparison}
\renewcommand{\arraystretch}{1.25}
\setlength{\tabcolsep}{4pt}
\adjustbox{max width=\textwidth}{%
\begin{tabular}{>{\raggedright\arraybackslash}p{3.7cm}ccccccccccc}
\toprule
\textbf{Capability} & \textbf{\ak} & \textbf{lm-eval} & \textbf{HELM} & \textbf{LangSmith} & \textbf{DeepEval} & \textbf{Braintrust} & \textbf{OAI Evals} & \textbf{Inspect} & \textbf{Phoenix} & \textbf{Promptfoo} & \textbf{PyRIT} \\
\midrule
\multicolumn{12}{l}{\textit{Evaluation pipeline}} \\
Academic benchmarks
& \checkmark & \checkmark\ (rich) & \checkmark\ (rich) & \ding{55} & \textasciitilde & \ding{55} & \checkmark\ (reg.) & \checkmark\ (200+) & \ding{55} & \textasciitilde & \textasciitilde\ (safety) \\
LLM-as-judge
& \checkmark & \ding{55} & \ding{55} & \checkmark & \checkmark & \checkmark & \checkmark\ (m-graded) & \checkmark & \checkmark & \checkmark & \checkmark \\
Judge over logged data
& \checkmark & \ding{55} & \ding{55} & \checkmark & \textasciitilde & \checkmark & \textasciitilde & \textasciitilde & \checkmark & \textasciitilde & \textasciitilde \\
Performance profiling
& \checkmark & \ding{55} & \checkmark & \checkmark\ (traces) & \ding{55} & \checkmark\ (traces) & \ding{55} & \textasciitilde & \textasciitilde\ (traces) & \textasciitilde\ (assert.) & \ding{55} \\
Common API (bench+judge+perf)
& \checkmark & \ding{55} & \ding{55} & \ding{55} & \textasciitilde & \textasciitilde & \textasciitilde & \textasciitilde & \textasciitilde & \textasciitilde & \ding{55} \\
Direction-aware comparison
& \checkmark & \textasciitilde & \textasciitilde & \textasciitilde & \textasciitilde & \textasciitilde & \ding{55} & \ding{55} & \ding{55} & \ding{55} & \ding{55} \\
Provenance / fingerprint
& \checkmark & \ding{55} & \ding{55} & \ding{55} & \ding{55} & \ding{55} & \ding{55} & \textasciitilde & \ding{55} & \textasciitilde & \ding{55} \\
Zero required deps (core)
& \checkmark & \ding{55} & \ding{55} & \ding{55} & \ding{55} & \ding{55} & \ding{55} & \ding{55} & \ding{55} & \ding{55} & \ding{55} \\
Standalone, non-SaaS
& \checkmark & \checkmark & \checkmark & \ding{55}\ (SaaS) & \checkmark & \ding{55}\ (SaaS) & \textasciitilde & \checkmark & \textasciitilde\ (platform) & \checkmark & \checkmark \\
\midrule
\multicolumn{12}{l}{\textit{Agents, RAG, and red teaming}} \\
Tool-call / trajectory metrics
& \checkmark & \ding{55} & \ding{55} & \checkmark & \checkmark & \textasciitilde & \ding{55} & \checkmark & \textasciitilde & \checkmark & \ding{55} \\
Turn-aware parallel-call scoring
& \checkmark & \ding{55} & \ding{55} & \ding{55} & \ding{55} & \ding{55} & \ding{55} & \ding{55} & \ding{55} & \ding{55} & \ding{55} \\
Verified (state-based) outcome
& \checkmark & \ding{55} & \ding{55} & \textasciitilde & \textasciitilde & \textasciitilde & \ding{55} & \checkmark & \ding{55} & \textasciitilde & \ding{55} \\
RAG retrieval + grounding metrics
& \checkmark & \ding{55} & \ding{55} & \checkmark & \checkmark & \checkmark & \ding{55} & \textasciitilde & \checkmark & \checkmark & \ding{55} \\
Red teaming / adversarial probing
& \textasciitilde & \ding{55} & \textasciitilde & \ding{55} & \textasciitilde & \ding{55} & \ding{55} & \textasciitilde & \ding{55} & \checkmark & \checkmark \\
\bottomrule
\end{tabular}
}
\renewcommand{\arraystretch}{1.0}

\end{table}

Beyond the table, \ak\ treats a deployed agent as another backend and a recorded trajectory as another \code{precomputed} sample, so an agent and its base model go through the same pipeline. Against red-teaming tools such as Promptfoo and PyRIT, \ak\ is narrower: its probes are basic, and it targets measurement and comparison, not attack generation.

\FloatBarrier
\section{Design Principles}
\label{sec:design}

\paragraph{Interface design.} \code{Adapter}, \code{Model}, \code{Annotator}, and \code{Metric} each expose a minimal interface; custom components use the same mechanism as built-ins. A plain \code{list[str]} $\rightarrow$ \code{list[str]} callable is a valid model. \code{SCENARIOS}, \code{ADAPTERS}, \code{METRICS}, and \code{ANNOTATORS} accept a subclass plus one decorator. Model backends are dispatched by a prefix table in \code{AutoModel.resolve()}. The core declares \code{dependencies = []}; optional extras add backends and heavy metrics, imported lazily. The agent layer, \code{agent:} backend, tool-call parser, retrieval and deterministic RAG metrics, and all Lexsi importers are stdlib-only.

\paragraph{Score semantics.} Every \code{Metric} subclass must declare \code{MAXIMIZE} or \code{MINIMIZE}; omitting it raises \code{TypeError} at class definition (\code{Metric.\_\_init\_subclass\_\_}). Grading, diffing, and leaderboards share that declaration, so lower latency is ranked as better. This picks the right winner \emph{per metric}. It does not make two runs comparable, which also needs matching data, prompts, judge and decoding settings, and metric versions, and trading off heterogeneous metrics remains a separate problem. A SHA-256 hash covers model identity (including the resolved checkpoint), data, adapter and annotator configuration, metrics, and run settings; a matching fingerprint lets the runner skip re-execution. Closures are hashed by their immutable captured values, agent and RAG fields enter only when set, and an \code{agent:} run records the resolved endpoint and digests request headers. The fingerprint identifies a \emph{configuration}: identical fingerprints are necessary but not sufficient for identical results. It is a configuration identifier, not a reproducibility certificate.

\paragraph{Evidence handling.} A metric with nothing to measure returns no score; an unparseable judge verdict is an error, not a zero; \code{unknown} and \code{not\_tested} results are excluded from means and counted in \code{RunResult.unscored}. Imported agent episodes carry per-field coverage, an ineligible metric names the missing field, and an oracle with no state to check returns \code{unknown}, not \code{failure}. Tool schemas sent to a backend without \code{Capability.TOOLS}, or images sent to a text-only model, raise instead of being dropped. When a remote server may have ignored a request field (\code{chat\_template\_kwargs}, \code{parallel\_tool\_calls}), the run records it, prints it, and caps the affected scores.

\FloatBarrier
\section{System Architecture}
\label{sec:architecture}

\begin{figure}[t]
\centering
\resizebox{\textwidth}{!}{%
\begin{tikzpicture}[font=\sffamily]
\akcard{inad}{0}{2.1}{akslate}{Input \& Adapter}{1}{4.2cm}{}{%
  {scenario, samples, CSV, JSONL, HF dataset or folder, AgentTune, BFCL},
  {adapters for chat, few-shot, RAG, and tool calling},
  {each \code{Sample} becomes model-ready requests, tools attached},
  {fork: the native Model stage, or lm-evaluation-harness}}
\akcard{model}{4.75}{2.1}{akblue}{Model}{2}{4.2cm}{}{%
  {local and hosted models, APIs, deployed agents, and callables},
  {capability check: tools and images refused, never silently dropped},
  {retries failures, times every call},
  {returns the output and its trace: tool-call turns, messages, contexts}}
\akcard{metan}{9.5}{2.1}{akteal}{Metric \& Annotator}{3}{4.2cm}{}{%
  {optional Annotator extracts the answer; raw output kept},
  {metrics for accuracy, judging, tool use, retrieval, and RAG},
  {every score carries its direction: higher or lower is better},
  {unknown and not-tested results stay out of the mean}}
\akcard{runcmp}{14.25}{2.1}{akgreen}{RunResult \& Comparison}{4}{4.2cm}{}{%
  {per-metric statistics with latency, size, tokens, cost},
  {two-run diff or $N$-model leaderboard, direction-aware},
  {improvement or regression, retention, significance},
  {saved with provenance, pushed to the Hub}}

\draw[akmain] (inad.east) -- (model.west);
\draw[akmain] (model.east) -- (metan.west) node[midway, above=1pt, aklbl, text=akink] {+ trace};
\draw[akmain] (metan.east) -- (runcmp.west);

\draw[densely dotted, ultra thick, akgray, -{Stealth[length=2mm]}]
    (model.north) -- ++(0,0.7) -| (metan.north);
\node[aklbl, text=akgray, anchor=south] at ($(model.north)!0.5!(metan.north)+(0,0.75)$)
    {\code{model="precomputed"}: replay stored outputs or a recorded agent trace instead of calling a model};

\akcard{lmharness}{0}{-3.0}{akgray}{lm-evaluation-harness}{H}{1.5cm}{dashed}{%
  {\code{ak.run\_lmeval()}: a real harness run, the same \code{RunResult} back}}
\akcard{agent}{9.5}{-3.0}{akviolet}{Agent episode layer}{A}{1.5cm}{}{%
  {verified outcome, coverage, reliability (Figure~\ref{fig:agent-architecture})}}
\akcard{fingerprint}{14.25}{-3.0}{akamber}{Fingerprint}{F}{1.5cm}{}{%
  {SHA-256 over model, data, adapter, metrics, config: the cache and comparison key}}

\draw[akbranch, akgray] (inad.south) -- (lmharness.north)
    node[midway, right=2pt, aklbl, text=akgray] {alternate engine};
\draw[akbranch, akgray] (lmharness.south) -- ++(0,-0.4) -| (11.875,-1.3) -- (runcmp.west |- 0,-1.3);
\draw[akmain, akviolet] (metan.south) -- (agent.north) node[midway, right=2pt, aklbl, text=akviolet] {same metrics};
\draw[akmain, akamber] (runcmp.south) -- (fingerprint.north) node[midway, right=2pt, aklbl, text=akamber] {stamped};
\end{tikzpicture}%
}
\caption{Evaluation flow through the shared result object, including replay, harness, agent, and fingerprint paths. Solid, dotted, and dashed arrows denote standard execution, replay, and harness execution, respectively.}
\label{fig:architecture}
\end{figure}
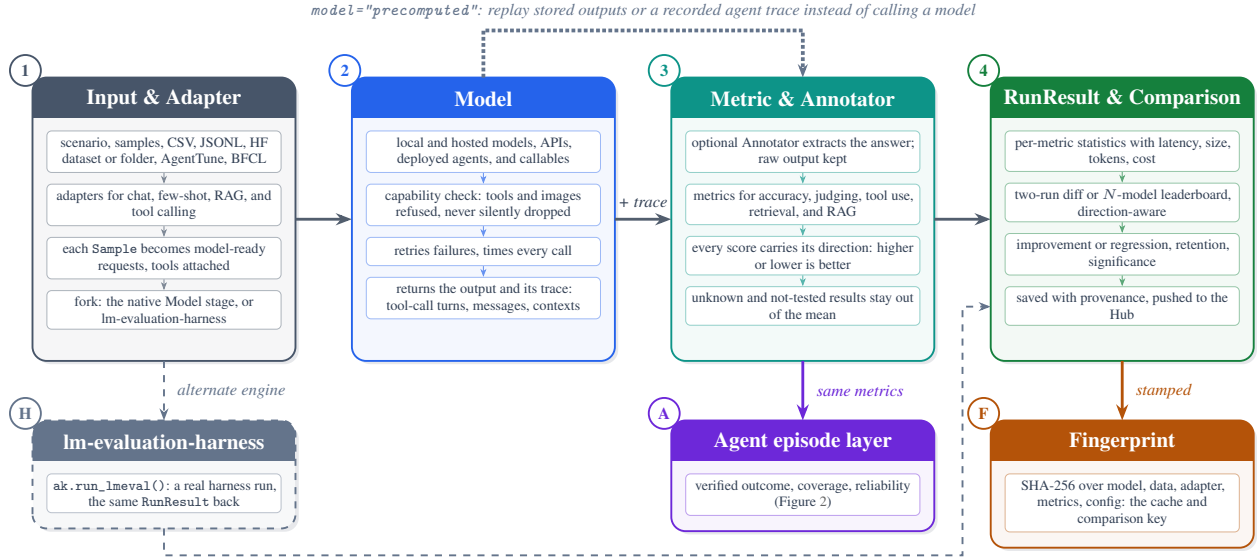

\ak\ is organized around an evaluation pipeline (Figure~\ref{fig:architecture}), a comparison layer that consumes its output, and an agent episode layer on top (Figure~\ref{fig:agent-architecture}, Section~\ref{sec:agentic}). Alongside each prediction, the pipeline carries a structured \emph{trace} (tool-call turns, messages, retrieved contexts) from backend to metrics through \code{context["trace"]}; for text-only evaluation it is empty.

\subsection{Evaluation pipeline}

\paragraph{Dataset $\rightarrow$ Samples.} Built-in scenarios (\code{mmlu}, \code{gsm8k}, \code{arc}, \code{hellaswag}, \code{truthfulqa}, \code{humaneval}), in-memory samples, CSV, JSONL, Hugging Face datasets or local folders (\code{ak.load\_dataset()}, including CuratorKIT exports), Croissant, AgentTune files (\code{ak.load\_agenttune()}), and BFCL (\code{ak.load\_bfcl()}) all become \code{Sample} objects: input, optional target, choices or retrieval context, metadata, and for agents and RAG the offered \code{tools}, reference \code{expected\_tool\_calls}, gold \code{reference\_contexts}, a recorded \code{actual\_trace}, and optional \code{images}.

\paragraph{Adapter $\rightarrow$ Requests.} An adapter (Table~\ref{tab:adapters}) turns each sample into requests, keeping the prompting strategy separate from data and model.

\paragraph{Model $\rightarrow$ Results.} \code{AutoModel.resolve("prefix:name")} dispatches to a registered backend or any callable. The backends cover local and hosted models, including vLLM~\cite{kwon2023vllm} and SGLang~\cite{zheng2024sglang}, OpenAI-compatible endpoints, and deployed agents. The runner retries transient failures, times every call, and copies the backend's trace (or a recorded \code{actual\_trace}) into the scoring context.

\paragraph{Annotator (optional).} Extracts a scoreable value from raw text without altering it (Listing~\ref{lst:annotator}).

\paragraph{Metric $\rightarrow$ Scores.} Each \code{Metric.score(sample, output, context)} returns \code{Score} objects stamped with the metric's direction.

\paragraph{Aggregate $\rightarrow$ RunResult.} Scores roll up into mean, standard deviation, standard error, and count; \code{RunResult} also holds predictions, fingerprint, configuration, errors, unscored counts, and measured performance, size, tokens, and cost.

\begin{lstlisting}[float=tbp, language=Python, caption={Basic evaluation: any dataset, any metric.}, label=lst:basic]
import auditkit as ak

samples = [
    ak.Sample(input="What is 2+2?", target="4"),
    ak.Sample(input="Capital of France?", target="Paris"),
]

# The test double `echo` needs no keys or downloads: zero-cost first contact.
result = ak.evaluate(samples, model="echo")
print(result.summary())
print(result.headline)   # {'exact_match': 0.0}
\end{lstlisting}

\subsection{Adapters and annotators}
\label{sec:components}

An adapter changes only how a sample becomes a request; Table~\ref{tab:adapters} lists the built-in adapters and Listing~\ref{lst:adapters} selects them. Annotators provide regex, reasoning-span, and model-based extraction; \code{extract\_with=} routes one annotator's output into scoring while \code{raw\_output} is kept.

\begin{table}[htbp]
\centering
\caption{Eight built-in adapters and their roles.}
\label{tab:adapters}
\adjustbox{max width=\textwidth}{
\begin{tabular}{l>{\raggedright\arraybackslash}p{3.6cm}>{\raggedright\arraybackslash}p{5.0cm}>{\raggedright\arraybackslash}p{5.2cm}}
\toprule
\textbf{Name} & \textbf{Class} & \textbf{Purpose} & \textbf{Representative parameters} \\
\midrule
\code{generation} & \code{GenerationAdapter} & Free-form generation (the default) & none \\
\code{mcq} & \code{MCQAdapter} & Multiple choice, using a sample's \code{choices} & \code{method="mcq\_joint"} or \code{"mcq\_loglikelihood"} \\
\code{chat} & \code{ChatAdapter} & Chat-style prompting with a system prompt & \code{system\_prompt="You are a helpful assistant."} \\
\code{fewshot} & \code{FewShotAdapter} & In-context few-shot examples & \code{num\_shots=3}, \code{separator="\textbackslash n\textbackslash n"}, \code{pool=None} \\
\code{instruction} & \code{InstructionAdapter} & Instruction-prefixed prompting & \code{instruction="Answer the following question:"} \\
\code{rag} & \code{RAGAdapter} & Retrieval-augmented generation using a sample's \code{retrieval\_context} & \code{context\_separator="\textbackslash nContext:\textbackslash n"}, \code{max\_context\_chars=None} \\
\code{template} & \code{TemplateAdapter} & Fully custom prompt template & \code{template="\{input\}"} \\
\code{tools} & \code{ToolCallAdapter} & Tool-use evaluation: sends a sample's \code{tools} as native schemas, or writes them into the prompt for text-only backends & \code{mode="native"} or \code{"prompt"}, \code{tool\_choice}, \code{parallel\_tool\_calls} \\
\bottomrule
\end{tabular}
}
\end{table}

\subsection{Model comparison and profiling}
\label{sec:comparison-layer}

Comparison consumes stored \code{RunResult} objects, so it works for every technique and custom metric. \code{compare\_models()} evaluates several models through the same pipeline and selects winners per metric by direction (Listing~\ref{lst:compare-models}); \code{RunComparison} adds \code{retention()}, \code{tradeoff()}, \code{regressed()}, and paired-bootstrap \code{significance()} (Listing~\ref{lst:diff}). Models need not share one \code{RunConfig}: \code{configs} and \code{model\_opts} override generation and backend settings per model (Listing~\ref{lst:compare-config}), and local models are loaded and freed one at a time.

\begin{lstlisting}[float=tbp, language=Python, caption={Comparing multiple models on the same metric set.}, label=lst:compare-models]
import auditkit as ak

results = ak.compare_models(
    ["openai:gpt-4o-mini", "anthropic:claude-haiku-4-5", "hf:Qwen/Qwen2.5-7B"],
    dataset=samples,
    scorers=["exact_match", "f1_score"],
)
print(results.summary())   # per-metric leaderboard, direction-correct winners
print(results.winner("exact_match"))
\end{lstlisting}

Each \code{RunResult} records latency, throughput, model size, token usage, and, given prices, cost (Listing~\ref{lst:perf}). Latency is sampled per \code{generate()} call, so a local \code{hf:}/\code{vllm:} run at \code{concurrency=1} that answers the whole dataset in one call records one sample: the run's wall-clock time, not per-request latency. API backends populate token usage; the local \code{hf:} backend does not.

\subsection{Model backends}
\label{sec:runtimes}

Native tool calling is available on \code{api:} (schemas, \code{tool\_choice}, and \code{parallel\_tool\_calls} forwarded; returned calls become one turn), \code{agent:}, and the local \code{hf:} and \code{vllm:} backends, which pass schemas to the chat template and parse calls from the reply. Any other backend is evaluated with \code{ToolCallAdapter(mode="prompt")}; the text parser reads Hermes/Qwen \code{<tool\_call>} blocks, JSON, and Cohere action formats in linear time. \code{RunConfig.chat\_template\_kwargs} (e.g.\ disabling a reasoning model's thinking) applies on \code{hf:}, \code{vllm:}, and \code{api:} and enters the fingerprint; \code{Sample.images} reach the model on \code{hf:} and \code{api:}; \code{hf:} also loads PEFT adapter folders. Study~4 reports measured coverage across runtimes; the versions are stated with Study~4.

\FloatBarrier
\section{Using AuditKIT}
\label{sec:usage}

The same \code{evaluate()} call serves every technique: changing the technique changes the components passed in, not the execution path. Listings~\ref{lst:swap-model} and~\ref{lst:judge} swap the backend and add an LLM judge; custom metrics (Listing~\ref{lst:custom-metric}) use the same path as built-in metrics.

\subsection{Academic benchmarks through \texttt{lm-evaluation-harness}}
\label{sec:lmeval}

\code{ak.run\_lmeval()} invokes the real \texttt{lm-evaluation-harness}~\cite{lmevalharness} and maps its results into the same \code{RunResult}, reusing the harness's task definitions while gaining \ak's provenance, caching, and comparison.

\begin{lstlisting}[float=tbp, language=Python, caption={Academic benchmarks via the real lm-evaluation-harness.}, label=lst:lmeval]
import auditkit as ak

r = ak.run_lmeval(
    ["arc_easy", "hellaswag", "gsm8k"],
    model="hf:meta-llama/Llama-3.2-1B",
    limit=100, device="cpu",
)
print(r.headline)
# {'arc_easy:acc': 0.61, 'hellaswag:acc_norm': 0.58, 'gsm8k:exact_match': 0.34, ...}
\end{lstlisting}

\subsection{Experiment management}

\code{evaluate\_many()} evaluates one model on several datasets (Listing~\ref{lst:many}); YAML and the CLI expose the same operations (Appendix~\ref{sec:appendix-usage-listings}). \code{ak.generate()} stores outputs once so that \code{model="precomputed"} can re-score them under other metrics, each scoring configuration receiving its own fingerprint. \code{ExperimentDB} keeps a queryable history, optionally mirrored to MLflow~\cite{mlflow}. When a model or dataset folder carries \code{lexsi\_provenance.json}, \ak\ records it in \code{RunResult.metadata["inputs"]}, writes it next to saved results, and uploads it with \code{push\_to\_hub}, so a result traces back to its checkpoint and data.

\FloatBarrier
\section{Metric Library, Custom Metrics, and Direction-Aware Comparison}
\label{sec:metrics}

\subsection{The metric contract and custom metrics}

Every metric implements \code{score(sample, output, context) -> Score} and declares a \code{direction}; omitting it raises \code{TypeError}. A function decorated with \code{@ak.scorer(direction=...)} (Listing~\ref{lst:custom-metric}) or a full \code{Metric} subclass (Listing~\ref{lst:full-metric}) is scored, aggregated, and compared exactly like a built-in.

\begin{lstlisting}[float=tbp, language=Python, caption={A custom metric, registered and used identically to a built-in metric.}, label=lst:custom-metric]
import auditkit as ak

# direction= is required: omitting it raises TypeError at class-definition
# time (Section 4.1), so a winner is never picked from an undeclared metric.
@ak.scorer(direction=ak.Direction.MAXIMIZE)
def mentions_price(sample, output) -> float:
    return 1.0 if "$" in output else 0.0

result = ak.evaluate(
    samples, model="openai:gpt-4o-mini",
    # Custom scorer and a built-in metric name side by side: same list, same code path.
    scorers=[mentions_price, "exact_match"],
)
\end{lstlisting}

\subsection{Supported metrics}

Table~\ref{tab:metrics} summarizes the built-in families. The generation family implements BLEU~\cite{bleu}, ROUGE~\cite{rouge}, and BERTScore~\cite{bertscore}; \code{GEval} implements G-Eval~\cite{geval}. The four lexical RAG metrics are named for the signals they compute, not after RAGAS~\cite{ragas}; the judged RAG family does claim-level judging and is kept separate so the two methodologies are never averaged under one name. The agent, retrieval, and deterministic RAG families are detailed in Section~\ref{sec:agentic} and Table~\ref{tab:agent-metrics}.

\needspace{30\baselineskip}
\small
\begin{longtable}{p{2.6cm}p{4.6cm}p{5.0cm}c}
\caption{Built-in metric families, representative metrics, and registry counts (registered implementations, not independently validated metrics).}
\label{tab:metrics}\\
\toprule
\textbf{Family} & \textbf{Representative metrics} & \textbf{What it measures} & \textbf{\#} \\
\midrule
\endfirsthead
\multicolumn{4}{l}{\small\itshape Table~\thetable{}, continued from previous page}\\
\toprule
\textbf{Family} & \textbf{Representative metrics} & \textbf{What it measures} & \textbf{\#} \\
\midrule
\endhead
\midrule
\multicolumn{4}{r}{\small\itshape continued on next page}\\
\endfoot
\bottomrule
\endlastfoot
Built-in
& \code{ExactMatch}, \code{QuasiExactMatch}, \code{Acc}, \code{AccNorm}
& Exact / normalized correctness against a target & 4 \\

Code / deterministic
& \code{Contains}, \code{Regex}, \code{IsJson}, \code{F1Score}, \code{Levenshtein}, \code{Equals}, \code{StartsWith}, \code{EndsWith}, \code{WordCount}
& Structural or exact string/JSON correctness & 9 \\

Generation quality
& \code{Bleu}, \code{RogueL} (registry name \code{rouge\_l}), \code{ChrF}~\cite{chrf}, \code{WordErrorRate}, \code{Perplexity}, \code{BertScore}
& N-gram overlap, fluency, LM confidence & 6 \\

Embedding similarity
& \code{CosineSimilarity}, \code{TokenOverlap}, \code{BM25Similarity}
& Semantic and lexical similarity & 3 \\

Retrieval-augmented generation (RAG)
& \code{LexicalGroundedness}, \code{ContextOverlap}, \code{ContextCoverage}, \code{AnswerOverlap}
& Groundedness and retrieval quality (lexical-overlap heuristic) & 4 \\

Hallucination
& \code{FactualConsistency}
& Groundedness of generated claims against source text & 1 \\

Toxicity / bias
& \code{ToxicityScore}, \code{RepresentationSkew}, \code{HateSpeechScore}
& Content-safety signals & 3 \\

Pairwise / preference
& \code{WinRate}, \code{EloScore}, \code{PreferenceAccuracy}
& Model-vs-model preference comparison & 3 \\

Security
& \code{KeywordDetector}
& Presence of unsafe keywords/patterns & 1 \\

LLM-as-judge
& \code{LLMJudge}, \code{GEval} (weighted rubric), \code{Factuality}, \code{ClosedQA}, \code{Relevance}, \code{BiasJudge}
& Rubric- or criteria-based judging by another LLM & 6 \\

Encoder-judge
& \code{EncoderJudge}, \code{FactualityEncoderJudge}, \code{SentimentEncoderJudge}
& Rubric-style judging by a lightweight encoder classifier, not a generative LLM & 3 \\

Guardrail
& \code{GuardJudge}
& Content-moderation verdict from a dedicated guard model & 1 \\

Agent / tool use
& \code{ToolCallF1}, \code{TrajectoryMatch}, \code{ParallelToolCalls}, \code{ToolCallValidity}, \code{RedundantToolCalls}, \code{AgentLoopDetection}, \code{ToolPermission}, \code{TaskCompletion}, \code{ToolSelectionJudge}
& Call correctness, turn structure, parallel-call behavior, schema validity, redundancy, loops, least privilege, judged completion & 9 \\

Retrieval \& judged RAG
& \code{RetrievalMetrics}, \code{Faithfulness}, \code{ContextPrecision}, \code{ContextRecall}, \code{AnswerRelevancy}, \code{ResponseGroundedness}, \code{Hallucination}, \code{ContextRelevance}
& Ranked retrieval quality; claim- and sentence-level grounding judged by an LLM, with and without a reference & 8 \\

Deterministic RAG checks
& \code{EvidenceSetRecall}, \code{Freshness}, \code{AclCompliance}, \code{CitationSupport}, \code{Abstention}, \code{NumericAccuracy}, \code{ContextRetention}
& Model-free checks of evidence completeness, freshness, access control, citations, numbers, and abstention against a versioned corpus & 7 \\

\midrule
\textbf{Total} & & & \textbf{68} \\
\end{longtable}
\normalsize

Performance measurements attach to every \code{RunResult} automatically. Some metrics emit more than one score, so registry size describes the available metrics rather than the number of result fields.

\FloatBarrier
\section{Evaluating Agentic and Retrieval-Augmented Systems}
\label{sec:agentic}

\ak\ evaluates agents at two levels that share one set of metric classes. At the \emph{metric level}, tool-calling models, deployed agents, and RAG pipelines go through \code{evaluate()} like any other task. At the \emph{episode level}, \code{auditkit.agent\_eval} evaluates a whole task with a verified outcome as the headline. Table~\ref{tab:agent-metrics} lists the metrics; the whole layer is stdlib-only and never imports AgentTune, \code{torch}, or \code{transformers}.

\begin{table}[t]
\centering
\caption{Agent and RAG metrics with their required inputs.}
\label{tab:agent-metrics}
\renewcommand{\arraystretch}{1.1}
\small
\adjustbox{max width=\textwidth}{%
\begin{tabular}{>{\raggedright\arraybackslash}p{5.2cm}>{\raggedright\arraybackslash}p{3.9cm}>{\raggedright\arraybackslash}p{6.4cm}}
\toprule
\textbf{Metric} & \textbf{Needs} & \textbf{Measures} \\
\midrule
\multicolumn{3}{l}{\textit{Agent / tool use}} \\
\code{tool\_call\_f1} & reference calls & precision, recall, F1, exact set; maximum bipartite matching \\
\code{trajectory\_match} & reference turns & agreement between the predicted and reference tool-call sequences \\
\code{parallel\_tool\_calls} & reference turns & independent calls batched (recall), dependent calls not (precision), when to batch (detection) \\
\code{tool\_call\_validity} & offered schemas & calls valid against the schemas \\
\code{redundant\_tool\_calls} $\downarrow$ & trace & exact repeats of an earlier call \\
\code{agent\_loop\_detection} & trace & repeated calls, stagnating messages, call-graph cycles \\
\code{tool\_permission} & allow / deny list & calls within a least-privilege policy \\
\code{task\_completion}, \code{tool\_selection} & judge model & judged completion; judged justification of each call \\
\midrule
\multicolumn{3}{l}{\textit{Retrieval and judged RAG}} \\
\code{retrieval}$(k)$ & gold contexts, ranked retrieval & hit rate, precision, recall, MRR, AP, nDCG \\
\code{faithfulness}, \code{context\_precision}, \code{context\_recall} & contexts (+ reference) & claims supported; relevant chunks ranked first; reference covered \\
\code{answer\_relevancy}, \code{response\_groundedness}, \code{hallucination} $\downarrow$, \code{context\_relevance} & question, contexts, answer; no reference & reference-free relevance and grounding \\
\midrule
\multicolumn{3}{l}{\textit{Deterministic RAG checks (model-free)}} \\
\code{evidence\_set\_recall}, \code{context\_retention} & case evidence sets, trace & a complete sufficient evidence set retrieved; still present in the final context \\
\code{freshness}, \code{acl\_compliance} & corpus snapshot, case date and identity & no superseded documents; no unauthorized or cross-tenant documents \\
\code{citation\_support}, \code{numeric\_accuracy}, \code{abstention} & gold citations, gold answer, answerability & cited spans exist and support; numbers correct; abstains exactly when it should \\
\bottomrule
\end{tabular}
}
\end{table}

\subsection{Trace representation}
\label{sec:agent-trace}

A reference is a list of \emph{turns}, each the set of calls made in one response: \code{[[a, b], [c]]} says \code{a} and \code{b} are independent and belong together, while \code{c} depends on them; \code{[]} means no tool should be called. Calls may be \code{\{name, arguments\}} records, OpenAI \code{tool\_calls}, or BFCL's form; arguments that do not decode mark the call as unmatched. The trace comes from native tool calls, a deployed \code{agent:} endpoint, a recorded run under \code{precomputed}, or, failing those, the parsed text (Listing~\ref{lst:agent-trace}). When a trace marks tool coverage unavailable, the reference metrics are ineligible rather than scored as ``no call''.

\subsection{Tool-call and trajectory metrics}
\label{sec:agent-metrics}

Let the reference be turns $R = (R_1, \dots, R_m)$ and the run $P = (P_1, \dots, P_n)$, and $\mathcal{C}(\cdot)$ the multiset of all calls. A predicate $\mu(p, r)$ requires equal tool names and, for arguments, equality (\code{"exact"}), reference-containment (\code{"subset"}: every reference argument must appear in the call, with extra call arguments allowed), or no argument check (\code{"name"}), or a custom \code{arg\_match}. Calls are paired by a maximum bipartite matching $M$ (Hungarian-style augmenting paths~\cite{kuhn1955hungarian}), so a greedy first fit never rejects a correct run. Precision is $|M|/|\mathcal{C}(P)|$ and recall $|M|/|\mathcal{C}(R)|$. With $G = \{R_i : |R_i| \geq 2\}$ the parallel groups and $Q$ the predicted turns holding at least two matched calls,
\begin{align*}
\text{par.\ recall} &= \tfrac{1}{|G|}\textstyle\sum_{g \in G} \mathbf{1}\big[\exists j:\ g \text{ fully matched in } P_j\big], \\
\text{par.\ precision} &= \tfrac{1}{|Q|}\big|\{P_j \in Q:\ \text{the matched calls of } P_j \text{ come from one } R_i\}\big|,
\end{align*}
and detection is $\mathbf{1}[(\exists j: |P_j| \geq 2) = (\exists i: |R_i| \geq 2)]$. Low recall means independent calls were serialized; low precision means dependent calls were batched before their inputs existed (Listing~\ref{lst:parallel}). These scores measure adherence to the reference's turn structure, not task success: serializing independent calls can be functionally correct, and a trace of calls does not show that their results were used. Scores without a denominator are omitted, not zeroed.

\begin{lstlisting}[float=tbp, language=Python, caption={Parallel-call scoring: three runs with identical call-level F1 are separated by turn structure. Real output.}, label=lst:parallel]
import auditkit as ak
W = lambda city: {"name": "get_weather", "arguments": {"city": city}}
BOOK = {"name": "book_flight", "arguments": {"flight_id": "AZ-61"}}
REF = [[W("Paris"), W("Rome")], [BOOK]]   # a parallel group, then a dependent call

runs = {"correct":    [[W("Paris"), W("Rome")], [BOOK]],
        "serialized": [[W("Paris")], [W("Rome")], [BOOK]],
        "batched":    [[W("Paris"), W("Rome"), BOOK]]}
for name, turns in runs.items():
    s = ak.Sample(input="Which is warmer, Paris or Rome? Book AZ-61 if it's Rome.",
                  expected_tool_calls=REF, actual_output="", actual_trace={"tool_calls": turns})
    h = ak.evaluate([s], model="precomputed",
                    scorers=[ak.ToolCallF1(), ak.ParallelToolCalls()]).headline
    print(name, h["tool_call_f1"], h["parallel_recall"], h.get("parallel_precision"))
# correct    1.0 1.0 1.0
# serialized 1.0 0.0 None    <- independent calls split across turns
# batched    1.0 1.0 0.0     <- booked before the weather result existed
\end{lstlisting}

A reference may list several valid routes, \code{\{"any\_of": [...]\}}. One route is chosen per sample by the key (F1, recall, exact turn structure, in-order coverage, listing order), and \emph{every} reference metric scores that same route, so a row never describes two different runs; scores record the chosen route and per-route F1, and \code{RunResult.route\_distribution()} reports which rule decided. Because the route is chosen with the prediction in view, these scores measure best-route reference compatibility, an optimistic measure, not independent evidence of task success. Reference-free checks need only the trace, and \code{ak.load\_bfcl()} with \code{bfcl\_arg\_match} reimplements BFCL's accepted-value rules. A non-default argument mode or custom matcher suffixes the score name, so differently configured instances are never averaged together.

\subsection{Task outcomes and reliability}
\label{sec:agent-episodes}

\begin{figure}[htbp]
\centering
\resizebox{\textwidth}{!}{%
\begin{tikzpicture}[font=\sffamily]
\akcard{src}{0}{2.1}{akslate}{Sources \& Importers}{1}{4.2cm}{}{%
  {AgentTune files and \code{EventLog}s},
  {OpenAI chat transcripts, a \code{Sample}'s recorded trace},
  {a deployed \code{agent:} endpoint, or a harness loop with tool doubles},
  {per-field coverage: observed, inferred, or unavailable}}
\akcard{ep}{4.75}{2.1}{akviolet}{AgentEpisode}{2}{4.2cm}{}{%
  {ordered events with role, type, call id, turn id},
  {final answer, final state snapshot, artifacts},
  {counters, stop reason, errors, latency},
  {the producer's own reward kept as provenance only}}
\akcard{orc}{9.5}{2.1}{akamber}{Outcome Oracle}{3}{4.2cm}{}{%
  {final state, artifact, answer, custom predicate, or a natural-language assertion},
  {verdict: success, failure, unknown, or error},
  {no evidence gives unknown, never failure},
  {a judge never overrides a verified verdict}}
\akcard{res}{14.25}{2.1}{akgreen}{AgentEvalResult}{4}{4.2cm}{}{%
  {per-case status: completed, target error, ineligible, and others},
  {verified success over all cases and over decided cases},
  {trials: pass@$k$, all-$k$, independence flag},
  {judge or source disagreement with the oracle recorded}}

\draw[akmain] (src.east) -- (ep.west);
\draw[akmain] (ep.east) -- (orc.west);
\draw[akmain] (orc.east) -- (res.west);

\akcard{lex}{4.75}{-3.0}{akgray}{Lexsi Evidence}{L}{1.9cm}{dashed}{%
  {AlignTune, SafeTune, CuratorKIT, CircuitKIT artifacts as \code{SourceEvidence}},
  {attached only through caller-named ids}}
\akcard{dia}{9.5}{-3.0}{akteal}{Diagnostic Scorers}{D}{1.9cm}{}{%
  {tool-call F1, trajectory, parallel calls, validity, retrieval, judge},
  {run only on observed fields; else name the missing field}}
\akcard{sc}{14.25}{-3.0}{akgray}{Sidecar \& Rescore}{S}{1.9cm}{}{%
  {\code{write\_sidecar()}: one record per AgentTune trajectory id},
  {\code{rescore()}: new scorers, no agent call}}

\draw[akbranch, akgray] (lex.north) -- (ep.south) node[midway, right=2pt, aklbl, text=akgray] {join};
\coordinate (epout) at ([yshift=-1.3cm]ep.east);
\coordinate (resin) at ([yshift=-1.3cm]res.west);
\draw[akmain, akteal] (epout) -- (7.125,0 |- epout) |- (dia.west);
\draw[akmain, akteal] (dia.east) -- (11.875,0 |- dia.east) |- (resin);
\draw[akmain] (res.south) -- (sc.north) node[midway, right=2pt, aklbl] {saved};
\end{tikzpicture}%
}
\caption{Episode-level agent evaluation separates verified outcomes from trace diagnostics. Dashed arrows mark caller-supplied artifact links.}
\label{fig:agent-architecture}
\end{figure}
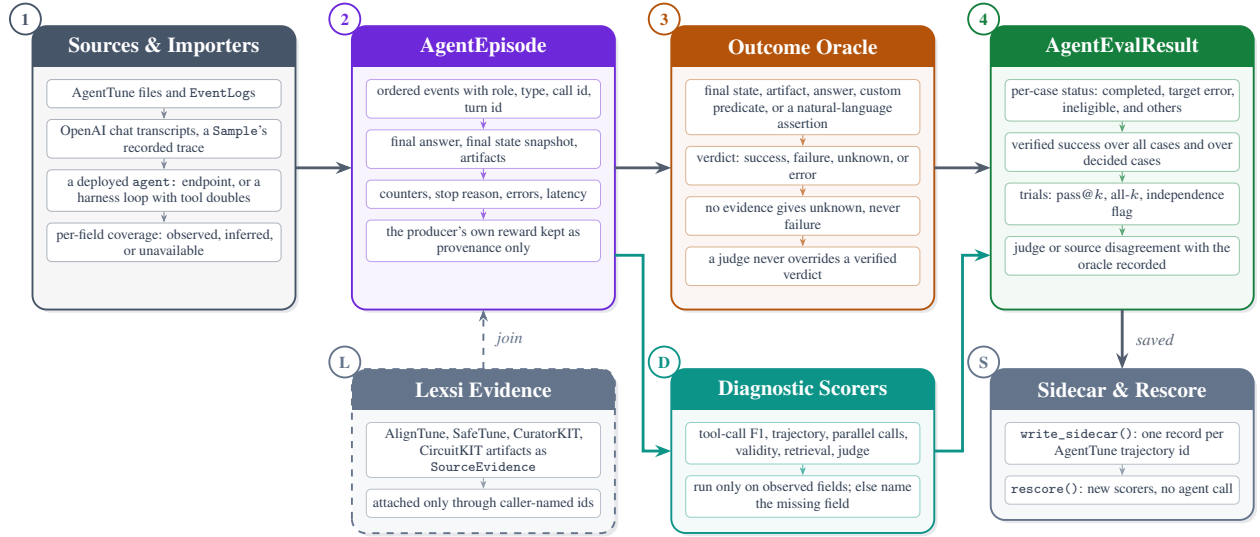

An \code{AgentCase} states the task, allowed tools, budgets, and an outcome oracle; an \code{AgentEpisode} is one run as ordered events (with call and turn ids) plus final answer, state, and artifacts (Figure~\ref{fig:agent-architecture}). Every imported episode reports, for nine fields, whether each was \code{observed}, \code{inferred}, or \code{unavailable}, and a scorer runs only when its fields were observed; otherwise the report names the missing field.

\paragraph{Oracles.} Oracles return \code{success}, \code{failure}, \code{unknown}, or \code{error}: \code{FinalStateAssertion} and \code{ArtifactAssertion} check a real effect, in the spirit of $\tau$-bench; \code{AnswerAssertion} checks the answer; \code{CustomPredicate} runs versioned caller code; and \code{AssertionOracle} turns a natural-language criterion into a verdict via a held-out judge, after an optional state check. Judges are useful but imperfect~\cite{zheng2023judging}, so \code{TaskCompletion} is a diagnostic that never overrides a verified oracle, and disagreements between judge and oracle are recorded. A verdict is relative to the specified oracle and the observed state, so its reliability depends on the oracle's correctness and coverage. A producer's own reward is kept as provenance only.

\paragraph{Execution modes.} Runs are \emph{recorded} (no calls), \emph{deployed} (an \code{agent:} endpoint under a wall-clock deadline), or \emph{harness} (\ak\ drives a policy against caller-supplied tool doubles). Each case ends in one of seven statuses; the summary gives success over all cases and over decided cases.

\paragraph{Reliability.} With $N$ trials, the headline is the majority of decided trials (a tie is \code{unknown}). The report also gives
\[
  \text{pass@}k = 1 - \binom{n-c}{k}\Big/\binom{n}{k}, \qquad \text{all-}k = \binom{c}{k}\Big/\binom{n}{k},
\]
where $\text{pass@}k$ follows Chen et al.~\cite{chen2021codex} and $\text{all-}k$ is the pass$^k$ of $\tau$-bench, flagged unless the caller confirms a reset between trials. \code{rescore()} replays saved episodes under new scorers without endpoint calls (Listing~\ref{lst:agent-eval}).

\begin{lstlisting}[float=tbp, language=Python, caption={Episode evaluation: a final-state oracle is the headline; tool scores are diagnostics; nothing is called.}, label=lst:agent-eval]
from auditkit.agent_eval import (AgentCase, AgentEvalRunner, AgentEvalSpec,
                                 FinalStateAssertion, episode_from_openai_messages)
episode = episode_from_openai_messages(messages, case_id="inventory-001",
                                       final_answer="Shipment for A is ready.")
episode.final_state = {"shipment_decision": "ready"}           # the verified effect
case = AgentCase(id="inventory-001", task="Ship item A if it is in stock.",
                 allowed_tools=["lookup_inventory"],
                 outcome=FinalStateAssertion("shipment_decision", equals="ready"))
result = AgentEvalRunner().run(AgentEvalSpec(
    cases=[case], mode="recorded", episodes=[episode], trials=1,
    scorers=["tool_call_f1", "tool_call_validity"]))   # add judge=... for task_completion
# Verified success: 1/1 (1/1 decided)   Failure: 0   Unknown outcome: 0 ...
\end{lstlisting}

\subsection{Retrieval-augmented generation}
\label{sec:rag}

A RAG pipeline is scored on what it actually retrieved: the trace's contexts when returned, the dataset's otherwise. \code{RetrievalMetrics(k)} is deterministic; its average precision divides by \emph{all} relevant items, so an unretrieved gold chunk lowers the score, and its ideal DCG~\cite{jarvelin2002ndcg} comes from the gold grades. Judged metrics work with any text judge and treat an unparseable verdict as an error; faithfulness fails unsupported claims, not only contradicted ones (Listing~\ref{lst:rag}).

In regulated settings such as banking, a grounded answer can still rest on a stale, unauthorized, or misattributed source. Seven model-free checks read a versioned \code{CorpusSnapshot}, a \code{RAGCase} (sufficient evidence sets, citations, tenant, decision date), and a \code{RAGTrace}, and each returns a score or an explicit \code{unknown}: a complete evidence set was retrieved and kept in the final context, the documents were current at the decision date, retrieval stayed within the caller's access rights, citations exist and support the answer, the numbers are right, and the system abstained exactly when the question was unanswerable.

\subsection{Integration with external artifacts}
\label{sec:agenttune}

\ak\ reads AgentTune~\cite{agenttune} artifacts by record shape, never by import, and sets coverage from what each artifact contains (Table~\ref{tab:agenttune}); a serialized \code{EventLog}'s turn boundaries become parallel groups, and AgentTune's healing-audit records, a per-call projection rather than a trajectory, are not imported. Several AgentTune metrics map onto \ak\ in stricter form: argument correctness to \code{tool\_call\_validity}, redundancy ratio to \code{redundant\_tool\_calls}, loop frequency to \code{agent\_loop\_detection}, tool-chain order to \code{trajectory\_in\_order}, and message-id recall to retrieval recall; self-correction, tool efficacy, plan adherence, and path edit distance have no counterpart yet. \ak\ never writes into AgentTune's data: \code{write\_sidecar()} emits verdicts, coverage, and reliability keyed by trajectory id (Listing~\ref{lst:agenttune}).

\begin{table}[htbp]
\centering
\caption{AgentTune artifact fields and the diagnostics available from them.}
\label{tab:agenttune}
\adjustbox{max width=\textwidth}{
\begin{tabular}{>{\raggedright\arraybackslash}p{3.6cm}>{\raggedright\arraybackslash}p{3.0cm}>{\raggedright\arraybackslash}p{5.4cm}>{\raggedright\arraybackslash}p{3.6cm}}
\toprule
\textbf{AgentTune artifact} & \textbf{\ak\ entry point} & \textbf{What is observed} & \textbf{Eligible without annotation} \\
\midrule
Evaluation / RL dataset rows (\code{prompt}, gold answer, gold chunk ids) & \code{load\_agenttune} & a task to run: input, target, gold contexts, and the original system prompt; no events yet & answer metrics; retrieval once a live run returns ids; \code{task\_completion} \\
Trajectory JSONL (\code{steps}) & \code{load\_agenttune}, \code{episodes\_from\_agenttune} & tool calls per step with arguments when recorded; conversation; retrieved chunk ids; reward as provenance & redundancy, loop detection, \code{task\_completion}; reference metrics once a reference is attached \\
Training trace (\code{trace.jsonl}) & same & final answer and retrieved ids; calls carry no tool names & answer metrics, \code{retrieval}, \code{task\_completion} \\
Evaluation report JSON & same & answers plus flattened tool \emph{names} only (\code{tool\_names\_only}); report metadata kept & \code{tool\_call\_f1\_name}, \code{tool\_call\_validity}, \code{task\_completion} \\
Serialized \code{EventLog} & \code{episode\_from\_eventlog} & ordered events, turn grouping (full tier), call ids, reward as provenance & all trajectory metrics the recorded tier supports \\
Healing audit records & refused (named error) & a lossy per-call projection & none \\
\bottomrule
\end{tabular}
}
\end{table}

The same read-only pattern covers AlignTune~\cite{aligntune} (scorecards and regression reports), SafeTune~\cite{safetune} (policy events and benchmark rows), CuratorKIT~\cite{curatorkit} (samples with provenance, manifests, trainer exports), and CircuitKIT~\cite{circuitkit} (circuit faithfulness reports). Each artifact is wrapped in a \code{SourceEvidence} envelope listing the join keys present and missing, and \code{join\_evidence()} attaches it to an episode only through ids the caller names, refusing to pin aggregate evidence to one case.

\section{Empirical Studies}
\label{sec:empirical}

Each study is a single run or a small or controlled sample (the sample size $N$, from 6 to 100, is stated with each). They illustrate what the library reports and distinguishes rather than general model rankings. Studies 1 to 3 ran on the earlier \texttt{auditkit} 0.3.0 line (44 metrics, 7 adapters, no agent layer, not published on PyPI); Studies 4 to 7 used the 1.1.2 release, Study 4 through its vLLM and SGLang integration code. All runs use greedy decoding, with output budgets (\code{max\_tokens}) of 8 for Study~1, 300 for Study~2, 100 for Study~3 (200 for its judge), 64 for Study~5, and 512 for Study~7. Studies 1 to 3 cover arbitrary metrics, direction-correct multi-model comparison, and performance measurement. Study 4 serves one model family through four runtimes. Studies 5 to 7 apply the RAG, interoperability, and tool-calling layers: a multi-hop question-answering assistant, artifacts from the Lexsi Labs stack, and the choice of a tool-calling model. Studies 5 and 6 are the smallest and illustrate what each layer reports.

\subsection{Study 1: Evaluating pruned models}
\label{sec:study1}
We evaluate three community-pruned MLP-neuron variants of \texttt{Llama-3.2-3B}~\cite{llama3} (the third-party \texttt{oopere/pruned\{20,40,60\}-llama-3.2-3b} checkpoints, whose pruning method and any post-pruning recovery are the checkpoint author's, not ours) against the unpruned base model on 100 BoolQ~\cite{boolq} questions, one run per severity and without repeated seeds. The baseline is the \emph{base} (not instruct) model with a one-shot prompt, and the answer is a \code{RegexAnnotator} match on \code{\textasciicircum\textbackslash s*(yes|no)}. Each variant is compared with the baseline through \code{RunComparison.retention()} (candidate $\div$ baseline) on three metrics: \code{boolq\_answer\_match} (exact-match answer scoring), \code{not\_degenerate\_output} (flags only three identical consecutive words; it does \emph{not} detect empty output), and \code{factual\_correctness\_judge} (\code{LLMJudge}-based).

\begin{table}[htbp]
\centering
\caption{Relative metric retention by pruning severity. Baseline absolute scores were not retained, so the ratios are reported as relative retention only.}
\label{tab:pruning}
\adjustbox{max width=\textwidth}{
\begin{tabular}{lccc}
\toprule
\textbf{Pruning severity} & \textbf{boolq\_answer\_match ($\uparrow$)} & \textbf{not\_degenerate\_output ($\uparrow$)} & \textbf{factual\_correctness\_judge ($\uparrow$)} \\
\midrule
20\% & 0.411 & 1.000 & 0.385 \\
40\% & 0.726 & 1.000 & 0.788 \\
60\% & 0.630 & 1.000 & 0.673 \\
\bottomrule
\end{tabular}
}
\end{table}

Retention is non-monotonic across severities, and every variant fails the predefined threshold. The flat non-degeneracy score is a coarse signal, because the output budget is short and the heuristic misses empty outputs. The baseline's absolute scores were not retained and may be near chance, so the ratios describe relative retention rather than a pruning effect. Model size falls with severity, which confirms the pruning took effect.

\subsection{Study 2: Comparing language models}
\label{sec:study2}
We run \code{compare\_models()} over three local Hugging Face checkpoints (\texttt{Llama-3.1-8B-Instruct}~\cite{llama3}, \texttt{Llama-2-13b-chat-hf}~\cite{llama2}, \texttt{Qwen/Qwen2.5-14B-Instruct}~\cite{qwen25}; via \code{HFGenModel}) on 100 SST-2~\cite{sst2} sentiment samples with \code{QuasiExactMatch}, reporting per-model accuracy, latency, throughput, and model size (Table~\ref{tab:compare-example}).

\begin{table}[htbp]
\centering
\caption{Model scores and whole-run time on SST-2; run time is a whole-run measurement, not per-request latency.}
\label{tab:compare-example}
\adjustbox{max width=\textwidth}{
\begin{tabular}{lcccc}
\toprule
\textbf{Model} & \textbf{quasi\_exact\_match ($\uparrow$)} & \textbf{Params} &
\textbf{Size (MiB)} & \textbf{Run time (ms, $\downarrow$)} \\
\midrule
llama-8b (\texttt{Llama-3.1-8B-Instruct}) & \textbf{0.930} & 8.03B &
15{,}316.5 & \textbf{6{,}601} \\
llama-13b (\texttt{Llama-2-13b-chat-hf}) & 0.580 & 13.02B &
24{,}825.8 & 32{,}388 \\
qwen-14b (\texttt{Qwen2.5-14B-Instruct}) & 0.900 & 14.77B &
28{,}171.6 & 106{,}512 \\
\bottomrule
\end{tabular}
}
\end{table}

Adding regex extraction changes the apparent ranking even though the generated outputs are unchanged: the model that scores lowest without extraction becomes competitive after its labels are parsed. This shows how answer formatting can masquerade as a model-quality difference.

Whole-run wall time is not per-request latency and varied substantially between runs, so it is not used for a speed ranking. Per-sample scores were not retained, so we reconstructed confidence intervals and significance tests from aggregate accuracies; the paired test that suits these repeated examples is unavailable. The post-extraction differences are not statistically significant on these aggregates, and equivalence of the models is not tested. The study shows that generation, annotation, scoring, and performance measurement can be varied independently under one common \code{RunResult}.

\begin{table}[htbp]
\centering
\caption{Scores after regex extraction; timings are whole-run measurements from a separate run.}
\label{tab:study2-fixed}
\adjustbox{max width=\textwidth}{
\begin{tabular}{lcccc}
\toprule
\textbf{Model} & \textbf{quasi\_exact\_match ($\uparrow$)} & \textbf{Std.\ dev.} &
\textbf{$n$} & \textbf{Run time (ms, $\downarrow$, this run)} \\
\midrule
llama-8b (\texttt{Llama-3.1-8B-Instruct}) & 0.930 (unchanged) & 0.2551 & 100 & \textbf{64{,}861} \\
llama-13b (\texttt{Llama-2-13b-chat-hf}) & 0.910 (was 0.580) & 0.2862 & 100 & 112{,}349 \\
qwen-14b (\texttt{Qwen2.5-14B-Instruct}) & \textbf{0.960} (was 0.900) & 0.1960 & 100 & 107{,}349 \\
\bottomrule
\end{tabular}
}
\end{table}

\subsection{Study 3: Financial question answering}
\label{sec:study3}
Financial-filing question answering needs the \emph{right number}, not a plausible sentence, so numeric correctness and semantic grounding fail differently. We run \code{compare\_models()} on 100 question/answer/context triplets from \texttt{virattt/financial-qa-10K}~\cite{financialqa10k}; the first 100 rows are all NVIDIA (NVDA), so this is a single-company case study. It compares \texttt{Llama-3.1-8B-Instruct}~\cite{llama3} (fp16) with \texttt{Llama-2-13b-chat-hf}~\cite{llama2}, judged by an independent \texttt{Meta-Llama-3-8B-Instruct}~\cite{llama3}. Four scorers run per sample: \code{NumericFactMatch} (a notebook-defined heuristic that checks whether the first extracted number equals the first reference number, and breaks on unit changes such as ``\$26.97 billion'' vs ``26{,}974 million''), \code{LexicalGroundedness} and \code{BertScore} (grounding and semantic similarity against the filing excerpt), and \code{factual\_correctness\_judge}.

The judge is in the same family as \texttt{llama-8b}, so a self-preference effect is possible. \code{Llama-2-13b-chat-hf} and the judge were loaded 4-bit quantized to fit beside the unquantized 8B model on one GPU. This confounds the 8B-vs-13B comparison, because quantization can affect accuracy and run time as well as size: \texttt{llama-13b}'s reported size, 6{,}675.8\,MiB, is smaller than \texttt{llama-8b}'s fp16 15{,}316.5\,MiB despite more parameters.

\begin{table}[htbp]
\centering
\caption{Metric scores on questions from NVIDIA's 2023 10-K filing. The 13B model and the judge ran 4-bit quantized, and \code{numeric\_fact\_match} compares only the first extracted number.}
\label{tab:finance}
\adjustbox{max width=\textwidth}{
\begin{tabular}{lcccc}
\toprule
\textbf{Model} & \textbf{numeric\_fact\_match ($\uparrow$)} & \textbf{lexical\_groundedness ($\uparrow$)} & \textbf{bert\_score ($\uparrow$)} & \textbf{factual\_correctness\_judge ($\uparrow$)} \\
\midrule
llama-8b (fp16) & 0.3800 & \textbf{0.6659} & \textbf{0.8685} & 0.8300 \\
llama-13b (4-bit) & \textbf{0.4100} & 0.5812 & 0.7936 & \textbf{0.8700} \\
\bottomrule
\end{tabular}
}
\end{table}

The metrics disagree: one model leads on grounding, while the quantized larger model leads on numeric matching and the judge score. The latter gaps overlap their uncertainty intervals, so they are not used for a ranking. The case shows why metric disagreement and evaluation setup, including quantization and judge family, should be reported together.

\subsection{Study 4: Evaluation across serving runtimes}
\label{sec:study6}
We run nine Cohere models~\cite{coheremodels} through four routes: \code{hf:}, offline \code{vllm:}, SGLang via \code{api:}, and the Hugging Face router via \code{api:}. Each model answers one factual question in five languages (PASS $\geq$ 4/5), on a 40\,GB A100 or 95\,GB RTX PRO 6000. The \code{vllm:} column ran vLLM~0.30.0 with \texttt{transformers}~5.16.1, the SGLang column SGLang~0.5.20, and the \code{hf:} column \texttt{transformers}~5.17, measured between 2026-09-30 and 2026-10-01; all use greedy decoding. Image tests in the \code{vllm:} column use \code{vllm serve} through \code{api:}; offline \code{vllm:} does not accept images. Superscripts mark Transformers backends, SGLang threshold reporting, or matching answers across routes.

\begin{table}[htbp]
\centering
\caption{Correct answers out of five language probes by model and serving route. These are compatibility probes, not estimates of multilingual performance.}
\label{tab:study6}
\adjustbox{max width=\textwidth}{
\begin{tabular}{llcccc}
\toprule
\textbf{Model} & \textbf{Architecture} & \textbf{\code{hf:}} & \textbf{\code{vllm:}} & \textbf{SGLang via \code{api:}} & \textbf{HF router via \code{api:}} \\
\midrule
Tiny Aya Global & \code{cohere2} & 4/5 & 4/5 & 3/5$^\star$ & text \\
Tiny Aya Fire & \code{cohere2} & 3/5 & 4/5 & 3/5$^\star$ & text \\
Tiny Aya Water & \code{cohere2} & 4/5 & 4/5 & 3/5$^\star$ & text \\
Tiny Aya Earth & \code{cohere2} & 3/5 & 4/5 & 3/5$^\star$ & text \\
North Micro Vision Instruct & \code{cohere\_compass} & 3/5 & 5/5, image & PASS$^{\S\ddagger}$, image & text only \\
Aya Expanse 8B & \code{cohere} & 5/5 & 5/5 & 5/5 & text \\
Aya Expanse 32B & \code{cohere} & 5/5 & 5/5 & 5/5 & text \\
Aya Vision 8B & \code{aya\_vision} & 5/5 & 5/5$^\dagger$, image & 5/5$^\ddagger$, image & text only \\
Aya Vision 32B & \code{aya\_vision} & 4/5 & 5/5$^\dagger$, image & PASS$^{\S\ddagger}$, image & text only \\
\bottomrule
\end{tabular}
}
\end{table}

A model can load across serving routes yet produce different evaluation outcomes. In one route, a special token left in the response causes an exact-match failure; another route strips it. This small probe checks runtime compatibility, not comparative model quality.

\subsection{Study 5: Retrieval and answer quality}
\label{sec:study-rag}
A RAG assistant can fail in its retriever or in its generator, and the two need separate measurements. We answer six HotpotQA~\cite{yang2018hotpotqa} distractor questions, each with ten candidate paragraphs of which two hold the supporting facts, using two retrievers, BM25~\cite{robertson2009bm25} and TF-IDF, that pass their top three paragraphs to the same generator (\texttt{Qwen2.5-0.5B-Instruct}). Each sample carries the retrieved paragraphs as \code{retrieval\_context} and the gold paragraphs as \code{reference\_contexts}; \code{RetrievalMetrics(k=3)} scores what was retrieved, and quasi-exact match and token F1 score the answer. The prompt places the passages before the question.

\begin{table}[htbp]
\centering
\small
\caption{Retriever and answer metrics for HotpotQA ($N\!=\!6$, $k\!=\!3$, one shared generator).}
\label{tab:study-rag}
\begin{tabular}{lccccccc}
\toprule
\textbf{Retriever} & \textbf{Hit@3} & \textbf{P@3} & \textbf{R@3} & \textbf{MRR} & \textbf{nDCG@3} & \textbf{QEM} & \textbf{F1} \\
\midrule
BM25 & 0.833 & \textbf{0.500} & \textbf{0.750} & 0.750 & \textbf{0.704} & 0.333 & \textbf{0.370} \\
TF-IDF & 0.833 & 0.389 & 0.583 & 0.750 & 0.602 & 0.333 & 0.333 \\
\bottomrule
\end{tabular}
\end{table}

The retrievers tie on hit rate, but ranking-sensitive measures reveal differences in the retrieved evidence. Answer scores also change when the prompt places the question before rather than after the passages, while retrieval scores stay fixed. Six questions illustrate the measures rather than rank the retrievers; the case shows why retrieval and generation need separate measures.

\subsection{Study 6: Evaluating artifacts from the Lexsi Labs stack}
\label{sec:study-interop}
We test interoperability using sample artifacts from AgentTune (two trajectories), CuratorKIT, AlignTune, SafeTune, and CircuitKIT. The records are placeholders, so this study evaluates parsing and linking behavior, not model quality. An artifact-based link is checked against an identifier it contains; a caller-based link uses an identifier supplied externally.

\begin{table}[htbp]
\centering
\small
\caption{Evidence-linking outcomes for a sample AgentTune episode (constructed placeholder artifacts).}
\label{tab:study-interop}
\begin{tabular}{llll}
\toprule
\textbf{Evidence} & \textbf{Granularity} & \textbf{Ids given} & \textbf{Result} \\
\midrule
AlignTune regression report & aggregate & artifact name, format & joined (artifact) \\
AlignTune audit report & aggregate & checkpoint digest & joined (caller) \\
CircuitKIT faithfulness report & aggregate & model digest & joined (caller) \\
SafeTune audit event & per event & episode id, call id & joined (caller) \\
SafeTune audit event & per event & decision only & refused: describes, does not identify \\
CuratorKIT manifest & aggregate & row id & refused: aggregate cannot pin one row \\
\bottomrule
\end{tabular}
\end{table}

Final-state assertions verify the recorded episodes, while missing tool arguments and rewards remain unscored rather than inferred. Evidence records are joined only when their identifiers support the link; descriptive labels and aggregate manifests are refused when they cannot identify an episode.

\subsection{Study 7: Choosing a tool-calling model}
\label{sec:study-agent}
An assistant that books, ships, or refunds must pick the right tool, fill its arguments, issue independent calls together, and recognize when no tool applies. A larger model costs more to serve, so we ask what the extra parameters buy. We compare \texttt{Qwen2.5-7B-Instruct} and \texttt{Qwen2.5-14B-Instruct}~\cite{qwen25} on the first 100 questions of five single-turn BFCL~\cite{patil2025bfcl} categories: one call (\code{simple}), choosing among tools (\code{multiple}), independent calls in one response (\code{parallel}, \code{parallel\_multiple}), and questions no tool answers (\code{irrelevance}). Both models run through \code{vllm:} with native tool schemas via \code{ToolCallAdapter}, and every call is scored with BFCL's accepted-value rules (\code{bfcl\_arg\_match}). Call F1 uses maximum bipartite matching; exact set requires every call and argument to match. Parallel precision and recall apply to categories with independent reference calls. Validity is averaged over questions with a tool call, redundancy is lower-is-better, and in \code{irrelevance} F1/exact-set measure correctly making no call.

\begin{table}[htbp]
\centering
\small
\caption{Tool-calling metrics across BFCL categories. Validity is averaged over questions with a tool call; parallel precision and recall apply only to categories with independent reference calls (n/a otherwise).}
\label{tab:study-agent}
\adjustbox{max width=\textwidth}{
\begin{tabular}{llcccccc}
\toprule
\textbf{Category} & \textbf{Model} & \textbf{Call F1} & \textbf{Exact set} & \textbf{Par.\ recall} & \textbf{Par.\ precision} & \textbf{Validity} & \textbf{Redundant} \\
\midrule
\multirow{2}{*}{\code{simple}} & 7B & 0.980 & 0.98 & n/a & n/a & 1.000 & 0.000 \\
 & 14B & 0.980 & 0.98 & n/a & n/a & 1.000 & 0.000 \\
\midrule
\multirow{2}{*}{\code{multiple}} & 7B & 0.937 & 0.93 & n/a & n/a & 1.000 & 0.000 \\
 & 14B & 0.948 & 0.93 & n/a & n/a & 1.000 & 0.000 \\
\midrule
\multirow{2}{*}{\code{parallel}} & 7B & 0.912 & 0.89 & 0.89 & 1.00 & 1.000 & 0.000 \\
 & 14B & 0.919 & 0.89 & 0.89 & 1.00 & 0.993 & 0.000 \\
\midrule
\multirow{2}{*}{\code{parallel\_multiple}} & 7B & 0.905 & 0.80 & 0.83 & 1.00 & 0.989 & 0.000 \\
 & 14B & 0.897 & 0.81 & 0.84 & 1.00 & 0.989 & 0.012 \\
\midrule
\multirow{2}{*}{\code{irrelevance}} & 7B & 0.590 & 0.59 & n/a & n/a & 0.732 & 0.000 \\
 & 14B & \textbf{0.710} & \textbf{0.71} & n/a & n/a & 0.828 & 0.017 \\
\bottomrule
\end{tabular}
}
\end{table}

The models perform similarly on categories that require a tool call, while the larger model more often declines irrelevant requests; uncertainty intervals overlap, so this difference is suggestive. On parallel calls, call-level F1 preserves partial credit that exact-set scoring hides, whereas parallel precision is uninformative with single-turn references. These results describe this benchmark slice.

\FloatBarrier
\section{Limitations and Ethics Statement}
\label{sec:limitations}

\ak\ is a measurement library, and several parts of it are deliberately modest in scope. Image inputs are supported without image-specific metrics, and simulated-user conversation, perturbation, and tabular evaluation are planned extensions. Agent evaluation scores a single-response model one turn at a time, relies on the caller to declare trial independence, treats alternative routes as whole routes, and validates tool calls against the common subset of JSON Schema. Latency is sampled per \code{generate()} call, so a single-call local run reports one whole-run wall-clock sample rather than per-request latency (Section~\ref{sec:comparison-layer}); time-to-first-token and inter-token latency are outside the current scope, and the local \code{hf:} backend does not record token usage. Red-teaming support is intentionally basic compared with Promptfoo~\cite{promptfoo}, PyRIT~\cite{pyrit}, and DeepTeam~\cite{deepteam}, and its probes are dual-use and intended for authorized evaluation of models the user may test (Section~\ref{sec:related}).

The case studies show what the library reports, and a few of their features are worth keeping in mind. They are small. They use a single judge (\texttt{Meta-Llama-3-8B-Instruct}, greedy-decoded). The library can use other supported backends as judges, but each study ran one judge configuration, so judged verdicts have not been validated against human annotation and may favor a candidate from the judge's own model family (Study~3); the judge prompt is in Appendix~\ref{sec:appendix-judge}. Significance statements in Study~2 use the reported accuracies (95\% Wilson intervals and a two-proportion test) because per-sample paired scores were not retained.

This paper contains no harmful content, adversarial prompts, or personally identifiable information, and the agent harness runs tools only through caller-supplied doubles. \ak\ is released under the Lexsi Labs Source Available License v1.2, which is not an OSI-approved open-source licence and conditions commercial use; see the licence for its terms.

\FloatBarrier
\section{Conclusion}
\label{sec:conclusion}

We presented \ak, a source-available library that evaluates language models and agents through one run and comparison interface. It combines direction-aware metrics, configuration tracking, and extensible components with structured assessment of tool use, retrieval, and task outcomes. For agents, it keeps turn-level execution information, separates trajectory diagnostics from oracle-based outcomes, and reports when the available evidence is insufficient for scoring.

The seven case studies illustrate how evaluation choices shape reported results. Answer extraction changed the apparent model ranking while the generated responses stayed the same, serving-runtime differences changed exact-match outcomes, and retrieval metrics separated retrievers that hit rate treated as tied. These observations show why evaluation context should be retained and why each metric should be read together with the evidence it was computed from.

\paragraph{Availability.} \ak\ is available with documentation, a CLI, and examples (\url{https://github.com/Lexsi-Labs/AuditKIT}) under the Lexsi Labs Source Available License (LSAL) v1.2.

\bibliographystyle{plain}
\bibliography{references}

\appendix

\clearpage
\section{Appendix: Code Listings for Library API and Usage (Section~\ref{sec:usage})}
\label{sec:appendix-usage-listings}

The body keeps six short listings; the remaining examples it refers to are collected here, grouped as the core API (Appendix~\ref{sec:appendix-core}), agents and RAG (Appendix~\ref{sec:appendix-agents}), the command line, YAML configuration, and the judge prompt used in Studies~1 and~3.

\subsection{Core API}
\label{sec:appendix-core}

\needspace{10\baselineskip}
\begin{lstlisting}[language=Python, caption={Swapping the model backend: nothing else changes.}, label=lst:swap-model]
ak.evaluate(samples, model="hf:meta-llama/Llama-3.1-8B-Instruct",
            hf_token="hf_...", device="cuda")   # local, gated: needs an accepted-license HF token
ak.evaluate(samples, model="hf:gpt2", device="cpu")           # local, ungated HuggingFace
ak.evaluate(samples, model=lambda prompts: [call_my_app(p) for p in prompts])  # any callable
\end{lstlisting}

\needspace{26\baselineskip}
\begin{lstlisting}[language=Python, caption={Selecting different adapters for different prompting strategies.}, label=lst:adapters]
import auditkit as ak

# Chat-style prompting with a system prompt
ak.evaluate(samples, model="openai:gpt-4o-mini",
            adapter=ak.ChatAdapter(system_prompt="You are a helpful assistant."))

# Few-shot, drawing examples from a pool
ak.evaluate(samples, model="openai:gpt-4o-mini",
            adapter=ak.FewShotAdapter(num_shots=3, pool=fewshot_pool))

# Retrieval-augmented generation: `adapter="auto"` routes on sample shape:
# a sample carrying `retrieval_context` is routed to RAGAdapter automatically
# (and one carrying `tools` to ToolCallAdapter).
rag_data = [ak.Sample(
    input="When was the Eiffel Tower completed?",
    target="1889",
    retrieval_context=["The Eiffel Tower was completed in 1889 for the World's Fair."],
)]
ak.evaluate(rag_data, model="openai:gpt-4o-mini", adapter="auto",
            scorers=["lexical_groundedness", "context_coverage", "context_overlap"])
\end{lstlisting}

\needspace{19\baselineskip}
\begin{lstlisting}[language=Python, caption={A regex annotator extracts a clean answer before scoring.}, label=lst:annotator]
import auditkit as ak
from auditkit.annotator import RegexAnnotator

# A model that reasons before answering:
# "Reasoning...\nFINAL ANSWER: 42"
answer = RegexAnnotator(r"FINAL ANSWER:\s*(.+)", group=1, name="answer")

result = ak.evaluate(
    samples, model=reasoning_model, scorers="exact_match",
    annotators=answer, extract_with="answer",
)
print(result.predictions[0].raw_output)     # unchanged: the full reasoning + answer
print(result.predictions[0].parsed_answer)  # "42": what was actually scored
\end{lstlisting}

\needspace{25\baselineskip}
\begin{lstlisting}[language=Python, caption={LLM-as-judge scoring with a custom weighted rubric.}, label=lst:judge]
import auditkit as ak

# No `target` needed: GEval judges the output against the rubric, not a gold answer.
qa = [
    ak.Sample(input="What is the derivative of x^2?"),
    ak.Sample(input="Explain gradient descent."),
]

# Each RubricItem is one scored dimension; weight controls its share of the rubric total.
rubric = [ak.RubricItem("Correctness", weight=2.0, description="factually right"),
          ak.RubricItem("Clarity", weight=1.0)]

result = ak.evaluate(
    qa, model="hf:meta-llama/Llama-3.1-8B-Instruct", hf_token="hf_...",
    # judge_model is a second, independent model call per sample: distinct
    # from the model under evaluation above, so it never grades itself.
    scorers=[ak.GEval(rubric, judge_model="hf:meta-llama/Meta-Llama-3-8B-Instruct",
                      judge_model_args={"hf_token": "hf_..."})],
)
\end{lstlisting}

\needspace{19\baselineskip}
\begin{lstlisting}[language=Python, caption={One model, several datasets, one call.}, label=lst:many]
import auditkit as ak

results = ak.evaluate_many(
    {
        "refusal_bench": (refusal_samples, [refusal_judge]),
        "over_refusal":  (over_ref_samples, [over_refusal_judge]),
        "gsm8k_local":   (gsm_samples, ["exact_match"]),
    },
    model="hf:meta-llama/Llama-3.1-8B-Instruct", hf_token="hf_...",
    on_error="skip",   # one bad dataset load doesn't lose the rest
)
for name, r in results.items():
    print(name, r.headline)
\end{lstlisting}

\needspace{19\baselineskip}
\begin{lstlisting}[language=Python, caption={Two-run comparison (before/after a model change).}, label=lst:diff]
import auditkit as ak

base = ak.evaluate(data, model="hf:my-base-model", scorers=["exact_match"])
pruned = ak.evaluate(data, model="hf:my-pruned-model", scorers=["exact_match"])

cmp = ak.compare(base, pruned)
print(cmp.summary())
# exact_match: 0.8200 -> 0.7900  delta=-0.0300 (-3.7% rel)  (n=500->500, std=...)

cmp.retention("exact_match")        # fraction of baseline quality kept
cmp.significance()                  # paired bootstrap, aligned by sample_id
cmp.tradeoff(metric="exact_match")  # quality vs. measured size/latency
cmp.regressed()                     # the exact samples that got worse
\end{lstlisting}

\needspace{23\baselineskip}
\begin{lstlisting}[language=Python, caption={Accessing automatically measured performance and model size: real output, \texttt{Llama-3.1-8B-Instruct}, 100 samples.}, label=lst:perf]
import auditkit as ak

r = ak.evaluate(qa, model="hf:meta-llama/Llama-3.1-8B-Instruct", hf_token="hf_...",
                config=RunConfig(track_performance=True), scorers=[quality])

r.perf["latency_ms"]
# {"count": 1, "mean": 65174.78, "min": 65174.78, "max": 65174.78,
#  "p50": 65174.78, "p95": 65174.78, "p99": 65174.78, "std": 0.0}

r.perf["throughput"]
# {"rps": 1.534, "total_requests": 100, "total_time_ms": 65174.78,
#  "output_tokens_per_sec": 0.0, "total_tokens_per_sec": 0.0}

r.model_size
# {"is_local": True, "total_params": 8030261248, "nonzero_params": 8030259962,
#  "sparsity": 1.6e-07, "size_mb": 15316.51}

\end{lstlisting}

\needspace{34\baselineskip}
\begin{lstlisting}[language=Python, caption={Per-model configuration overrides in a single \texttt{compare\_models()} call.}, label=lst:compare-config]
import auditkit as ak
from auditkit.runspec import RunConfig

results = ak.compare_models(
    ["hf:base-model", "hf:pruned-model", "hf:meta-llama/Llama-3.1-8B-Instruct"],
    dataset=samples,
    scorers=["exact_match", "f1_score"],
    model_names=["base", "pruned", "prod"],

    # Applied to any model without its own entry below:
    config=RunConfig(temperature=0.0, max_tokens=256),

    # Per-model generation-level overrides (RunConfig), keyed by model_names:
    configs={
        "pruned": RunConfig(temperature=0.0, max_tokens=256, seed=7),
        "prod":   RunConfig(temperature=0.2, max_tokens=512),
    },

    # Per-model backend-construction-level overrides, keyed by model_names:
    model_opts={
        "base":   {"device": "cuda"},
        "pruned": {"device": "cuda"},
        "prod":   {"device": "cuda", "hf_token": "hf_..."},   # gated model
    },
)
print(results.summary())          # leaderboard + per-task tables + pairwise significance
print(results.winner("exact_match"))
cmp = results.pairwise("base", "pruned")   # zoom into any two -> RunComparison
\end{lstlisting}

\clearpage
\subsection{The full \texttt{Metric}-subclass form}

For scoring logic that needs constructor parameters or internal state, a metric can be a full \code{Metric} subclass; the registry treats both forms identically.

\needspace{29\baselineskip}
\begin{lstlisting}[language=Python, caption={A configurable custom metric, registered as a full \texttt{Metric} subclass.}, label=lst:full-metric]
from auditkit.metric import Metric
from auditkit.score import Score
from auditkit.types import Direction, ScoreKind
from auditkit.registry import METRICS

@METRICS.register("under_limit")  # registers the class under this string name
class UnderLimit(Metric):
    direction = Direction.MAXIMIZE  # class-level; still mandatory as with the decorator form

    def __init__(self, max_len=280):
        self.max_len = max_len
        self.name = f"under_limit({max_len})"  # per-instance name: distinguishes configs in results/fingerprint

    def score(self, sample, output, context=None):
        return Score(
            name=self.name,
            value=float(len(output) <= self.max_len),  # 1.0/0.0, not a raw length: direction stays meaningful
            kind=ScoreKind.CODE,
        )

# Used exactly like a built-in metric, by its registered name:
import auditkit as ak
result = ak.evaluate(samples, model="openai:gpt-4o-mini", scorers=["under_limit"])
\end{lstlisting}

\subsection{Agents and RAG}
\label{sec:appendix-agents}

\needspace{26\baselineskip}
\begin{lstlisting}[language=Python, caption={One tool-use sample through three trace sources: native tool calls from an OpenAI-compatible server, a deployed agent, and a recorded run.}, label=lst:agent-trace]
import auditkit as ak
TOOLS = [...]   # OpenAI-format schemas: get_weather, book_flight, compare_weather
W = lambda city: {"name": "get_weather", "arguments": {"city": city}}
BOOK = {"name": "book_flight", "arguments": {"flight_id": "AZ-61"}}
Q = "Which is warmer, Paris or Rome? Book AZ-61 if it's Rome."

sample = ak.Sample(input=Q, tools=TOOLS,
                   expected_tool_calls=[[W("Paris"), W("Rome")], [BOOK]])
scorers = [ak.ToolCallF1(), ak.TrajectoryMatch(), ak.ParallelToolCalls()]

# (i) vLLM / SGLang / OpenAI-compatible server; one request shows one turn
ak.evaluate([ak.Sample(input=Q, tools=TOOLS, expected_tool_calls=[[W("Paris"), W("Rome")]])],
            model="api:Qwen/Qwen2.5-7B-Instruct", api_base="http://localhost:8000/v1",
            adapter=ak.ToolCallAdapter(), scorers=scorers)
# (ii) a deployed agent that runs its own loop and returns its transcript
ak.evaluate([sample], model="agent:http://localhost:8000/run",
            adapter=ak.ToolCallAdapter(), scorers=scorers)
# (iii) a recorded run: no model is called
sample.actual_trace = {"tool_calls": [[W("Paris"), W("Rome")], [BOOK]]}
ak.evaluate([sample], model="precomputed", scorers=scorers)
\end{lstlisting}

\needspace{28\baselineskip}
\begin{lstlisting}[language=Python, caption={Alternative routes, reference-free trajectory checks, and BFCL data. Real output except the BFCL line, which needs the dataset download.}, label=lst:routes]
CMP = {"name": "compare_weather", "arguments": {"cities": ["Paris", "Rome"]}}
s = ak.Sample(input="Which is warmer, Paris or Rome?", tools=TOOLS,
              expected_tool_calls={"any_of": [[[W("Paris"), W("Rome")]],   # route 0
                                              [[CMP]]]},                   # route 1
              actual_output="", actual_trace={"tool_calls": [[CMP]]})
r = ak.evaluate([s], model="precomputed", scorers=[ak.ToolCallF1()])
# tool_call_f1 = 1.0, matched_path = 1, path_f1 = [0.0, 1.0]

# No reference needed: validity, redundancy, loops, least privilege
looping = ak.Sample(input=Q, tools=TOOLS, metadata={"allowed_tools": ["get_weather"]},
                    actual_output="", actual_trace={"tool_calls":
                        [[W("Paris")], [W("Paris")], [W("Paris")], [BOOK]]})
ak.evaluate([looping], model="precomputed", scorers=[
    ak.ToolCallValidity(), ak.RedundantToolCalls(), ak.AgentLoopDetection(),
    ak.ToolPermission(denied_tools=["book_flight"])]).headline
# {'tool_call_validity': 1.0, 'redundant_tool_calls': 0.5,
#  'agent_loop_detection': 0.0, 'tool_permission': 0.75}

# BFCL, scored with BFCL's own accepted-value rules
bfcl = ak.load_bfcl("BFCL_v3_simple.json", limit=100)
ak.evaluate(bfcl, model="api:Qwen/Qwen2.5-7B-Instruct", adapter=ak.ToolCallAdapter(),
            scorers=[ak.ToolCallF1(arg_match=ak.bfcl_arg_match)])
\end{lstlisting}

\needspace{28\baselineskip}
\begin{lstlisting}[language=Python, caption={RAG: ranked retrieval (real output), LLM-judged grounding, and deterministic evidence checks (real output).}, label=lst:rag]
s = ak.Sample(input="How often are laptops replaced?", target="Every 3 years.",
              reference_contexts=["it-04"], actual_output="Every 3 years.",
              actual_trace={"retrieved_contexts": ["it-09", "it-04", "hr-31"]})
ak.evaluate([s], model="precomputed", scorers=[ak.RetrievalMetrics(k=3)]).headline
# hit_rate@3 1.0  precision@3 0.333  recall@3 1.0  mrr@3 0.5  ap@3 0.5  ndcg@3 0.631

judge = dict(judge_model="api:Qwen/Qwen2.5-7B-Instruct",       # any text-in/text-out model
             judge_model_args={"api_base": "http://localhost:8000/v1", "api_key": "EMPTY"})
ak.evaluate(rag_samples, model="agent:http://localhost:8000/rag",   # scored on what it retrieved
            scorers=[ak.Faithfulness(**judge), ak.ContextRecall(**judge),
                     ak.Hallucination(**judge)])                      # hallucination: lower is better

# Deterministic checks: `snapshot` is a versioned policy corpus, `case` a question with its gold evidence
trace = ak.RAGTrace(query=case.question, retrieved=[{"id": "LP-CURRENT", "score": 0.9},
                                                    {"id": "FIN-TABLE", "score": 0.8}])
s = ak.Sample(input=case.question, actual_output="LTV cap is 80 percent; Q1 NII 1200.",
              actual_trace=trace.to_dict(),
              metadata={"rag_case": case.to_dict(), "corpus": snapshot.to_dict()})
ak.evaluate([s], model="precomputed",
            scorers=["evidence_set_recall", "freshness", "acl_compliance"]).headline
# {'evidence_set_recall': 1.0, 'freshness': 1.0, 'stale_hit_rate': 0.0,
#  'acl_compliance': 1.0, 'unauthorized_hit_rate': 0.0}
\end{lstlisting}

\needspace{22\baselineskip}
\begin{lstlisting}[language=Python, caption={A constructed AgentTune evaluation report (\code{report\_constructed.json}), re-scored as episodes and returned as a sidecar. Output on that constructed fixture.}, label=lst:agenttune]
import auditkit as ak
from auditkit.agent_eval import AgentEvalRunner, AgentEvalSpec, inspect_agenttune, write_sidecar

inspect_agenttune("report.json")["coverage_labels"]
# {'tool_names_only': 1, 'none': 1}   <- the report kept tool names, not arguments

episodes = ak.episodes_from_agenttune("report.json")
result = AgentEvalRunner().run(AgentEvalSpec(
    mode="recorded", episodes=episodes,
    scorers=["tool_call_f1_name", "tool_call_validity"]))
print(result.summary())
# Verified success: 1/2 (1/2 decided)   Failure: 1   Unknown outcome: 0 ...
result.rows[0].ineligible
# {'tool_call_f1_name': 'case.reference_turns', 'tool_call_validity': 'case.allowed_tools'}

write_sidecar(result, "auditkit_sidecar.jsonl")   # one record per trajectory id
\end{lstlisting}

\subsection{Command-line interface}

\needspace{8\baselineskip}
\begin{lstlisting}[language=bash, caption={CLI: zero-configuration smoke test.}, label=lst:cli-smoke]
pip install auditkit
auditkit eval --model echo <<< "What is 2+2?"  # zero-cost test double, no API key needed
\end{lstlisting}

\needspace{8\baselineskip}
\begin{lstlisting}[language=bash, caption={CLI: evaluating a built-in benchmark.}, label=lst:cli-eval]
auditkit eval --dataset mmlu --model openai:gpt-4o --output results.json
auditkit eval --dataset gsm8k --model openai:gpt-4o --output results.json
\end{lstlisting}

\needspace{9\baselineskip}
\begin{lstlisting}[language=bash, caption={CLI: evaluating against a CSV dataset: any tabular data, any column names.}, label=lst:cli-csv]
export HF_TOKEN=hf_...   # gated model: CLI has no --hf-token flag, reads the env var
auditkit eval --model hf:meta-llama/Llama-3.1-8B-Instruct --csv data.csv \
              --input-col question --target-col answer -o results.json
\end{lstlisting}

\needspace{8\baselineskip}
\begin{lstlisting}[language=bash, caption={CLI: comparing models from the command line.}, label=lst:cli-compare]
# --baseline picks which of --models is the reference for retention()/tradeoff().
auditkit compare --models "hf:base,hf:pruned" --dataset gsm8k --baseline hf:base
\end{lstlisting}

\needspace{8\baselineskip}
\begin{lstlisting}[language=bash, caption={CLI: listing every registered resource, or a single category.}, label=lst:cli-list]
auditkit list metrics
auditkit list all
\end{lstlisting}

\needspace{7\baselineskip}
\begin{lstlisting}[language=bash, caption={CLI: red-teaming a model with adversarial probes.}, label=lst:cli-redteam]
auditkit redteam --model echo --probes prompt_injection,jailbreak --detectors keyword
\end{lstlisting}

\needspace{19\baselineskip}
\begin{lstlisting}[language=bash, caption={CLI: agent evaluation. \texttt{-{}-inspect} and \texttt{-{}-dry-run} are local and read-only: they load no model and call no endpoint.}, label=lst:cli-agent]
# What an AgentTune file records, and which metrics it can feed
auditkit agent import-agenttune trajectories.jsonl --inspect
# Detected format(s): {'agenttune_trajectory': 1}
# Coverage labels: {'full': 1}
# Eligible metrics (trace-supported): answer, parallel_tool_calls, ...

# Validate the cases, oracles, trials and judge, and print the call plan
auditkit agent eval --config agent-suite.json --dry-run
# Run it, redacting a secret by the value of an environment variable
auditkit agent eval --config agent-suite.json --output agent-run.json \
                    --redact-env AGENT_API_KEY
# Rescore the saved episodes with other scorers: no agent call
auditkit agent rescore --run agent-run.json --scorers tool_call_f1,task_completion
\end{lstlisting}

\needspace{22\baselineskip}
\subsection{Configuration via YAML}

\begin{lstlisting}[language=bash, caption={A YAML configuration file, adapted from the library's own example configuration, run with \code{auditkit eval -{}-config auditkit.yaml}.}, label=lst:yaml]
# auditkit.yaml
model: openai:gpt-4o-mini
temperature: 0.0
max_tokens: 256
seed: 42
concurrency: 8

prompts:
  - "What is the capital of France?"
  - "Explain quantum computing in one paragraph."
  - "Write a haiku about programming."

adapter: generation
output: results.json
format: json

experiment: my-eval
tags:
  - test
  - v1
\end{lstlisting}

\subsection{Judge prompt}
\label{sec:appendix-judge}

The \code{factual\_correctness\_judge} prompt is identical in structure across Studies~1 and~3, differing only in the field naming the retrieved text (\code{Passage} for Study~1's BoolQ context, \code{Filing excerpt} for Study~3's 10-K excerpt):
\begin{quote}
\ttfamily\small
\{Passage/Filing excerpt\}: \{context\}\\
Question: \{input\}\\
Real answer: \{expected\}\\
Model answer: \{output\}\\
Does the model answer \{match/state the same fact as\} the real answer? Reply CHOICE: yes or CHOICE: no.
\end{quote}

\end{document}